\documentclass[sigconf]{Styles/acmart}

\usepackage{balance}

\usepackage[ruled,linesnumbered]{algorithm2e}

\SetAlgoNoLine

\usepackage{amsmath,amsfonts,bm}

\def\eqref#1{equation~\ref{#1}}

\def\1{\bm{1}}

\DeclareMathAlphabet{\mathsfit}{\encodingdefault}{\sfdefault}{m}{sl}
\SetMathAlphabet{\mathsfit}{bold}{\encodingdefault}{\sfdefault}{bx}{n}

\newcommand{\mytitle}{Rethinking Personalized Reward Modeling for LLMs under Preference
Heterogeneity via Group-Debiased Federated Learning}

\newcommand{\alg}{\textbf{\code{FedGD}}\xspace}
\newcommand{\algft}{\textbf{\code{FedGD-FT}}\xspace}

\newcommand{\acc}[2]{#1_{\pm #2}}   

\makeatletter
\renewcommand\@fnsymbol[1]{%
  \ifcase#1\or $\dagger$\or $\ddagger$\or $\S$\or $\P$\or $\|$\or
  $\ast\ast$\or $\dagger\dagger$\or $\ddagger\ddagger$\else\@ctrerr\fi}
\makeatother

\usepackage{booktabs}
\usepackage{makecell}
\usepackage{bbm}
\usepackage{mathtools}
\usepackage{nccmath}
\usepackage{setspace}
\usepackage[T1]{fontenc}
\usepackage[scaled]{beramono}
\usepackage{subcaption}

\usepackage[ruled,linesnumbered]{algorithm2e} %

\usepackage{tikz}
\definecolor{petalcolor}{HTML}{F5F0E6}
\definecolor{petaledge}{HTML}{2B2B2B}
\definecolor{corecolor}{HTML}{F2B705}
\definecolor{coredeep}{HTML}{C98A00}
\definecolor{corelight}{HTML}{FFD95C}

\SetKwInput{KwInput}{Input}                
\SetKwInput{KwOutput}{Output}              

\SetCommentSty{mycommfont}

\SetAlCapSty{algcapsty}

\usepackage[T1]{fontenc}
\usepackage{wrapfig,lipsum,booktabs}

\usepackage{soul}
\usepackage{dsfont}
\usepackage{enumitem}

\usepackage{amsmath}
\usepackage{amsfonts}
\usepackage{bbm}
\usepackage{dsfont}
\usepackage[Symbol]{upgreek}
\usepackage{lscape}
\usepackage{caption}
\usepackage{balance}
\usepackage{xspace}
\usepackage{float}

\usepackage{lipsum}

\usepackage{wasysym}
\usepackage{multirow}
\usepackage{array, boldline, rotating}
\usepackage[table]{xcolor}   
\definecolor{flhl}{RGB}{219,234,254}  
\definecolor{fthl}{RGB}{253,230,203}   

\usepackage{amssymb}
\usepackage{pifont}
\definecolor{LightCyan}{rgb}{0.88,1,1}
\definecolor{Blue}{rgb}{0, 0.5, 1}
\definecolor{Green}{rgb}{0.0, 0.8, 0.0 }
\definecolor{Red}{rgb}{0.95, 0.55, 0.6}
\definecolor{Skyblue}{rgb}{0.6, 0.6, 0.95 }

\renewcommand*\eqref[1]{(\ref{#1})}

\def\code#1{\texttt{#1}}

\NewDocumentCommand{\supptitle}{s}{
\onecolumn
\begin{center}
    \rule{\textwidth}{0.03cm}\\[0.1cm]
    - Appendix -\\[0.2cm]
    {\Large 
        \textbf{\mytitle }
    }\\
    \rule{\textwidth}{0.03cm}\\[0.2cm]
\end{center}
}

\AtBeginDocument{%
  }

\renewcommand\footnotetextcopyrightpermission[1]{}

\begin{document}


\title{Rethinking Personalized Reward Modeling for LLMs under Preference Heterogeneity via Group-Debiased Federated Learning
}



\author{Seongyoon Kim}
\affiliation{
  \institution{Institute of Engineering Research, Korea University}
  \city{Seoul}
  \country{Republic of Korea}
  \postcode{02841}
}
\email{curisam@korea.ac.kr}

\author{Boryeong Cho}
\affiliation{%
  \institution{Kim Jaechul Graduate School of AI, KAIST}
  \city{Seoul}
  \country{Republic of Korea}
  \postcode{02455}
}
\email{venntum@kaist.ac.kr}

\author{Jihwan Oh}
\affiliation{%
  \institution{Kim Jaechul Graduate School of AI, KAIST}
  \city{Seoul}
  \country{Republic of Korea}
  \postcode{02455}
}
\email{ericoh929@kaist.ac.kr}

\author{Seokhyun Chung}
\authornote{Corresponding authors.}
\affiliation{
  \institution{Industrial Management Engineering, Korea University}
  \city{Seoul}
  \country{Republic of Korea}
  \postcode{02841}
}
\email{csh8901@korea.ac.kr}

\author{Se-Young Yun}
\authornotemark[1]
\affiliation{%
  \institution{Kim Jaechul Graduate School of AI, KAIST}
  \city{Seoul}
  \country{Republic of Korea}
  \postcode{02455}
}
\email{yunseyoung@kaist.ac.kr}



\renewcommand{\shortauthors}{Seongyoon Kim et al.}


\begin{abstract}

Large language models are increasingly aligned to human preferences via reward modeling, but user preference data are sensitive and often cannot be centralized.
Federated learning keeps such data local while learning a shared initial reward model, which is later personalized for each client through local fine-tuning.
Because users often assign opposite labels to the same pair of responses, existing federated methods address preference heterogeneity by clustering similar clients and training one reward model per group, assuming that each group requires its own initialization. 
We show that this assumption is unnecessary. 
Under balanced preference groups, a single FedAvg model, despite starting at nearly random accuracy, surpasses reward models trained separately for each ground-truth group after only a few local optimization steps. 
We attribute this phenomenon to the flatness of the shared initialization: averaging across all clients learns richer shared representations that distinguish responses while canceling conflicting preference directions, leaving the model near a decision boundary that can be rapidly adapted.
Group imbalance breaks this effect as the cancellation becomes asymmetric and leaves minority clients too far from the boundary to recover. 
Motivated by this observation, we propose \alg\ (\textbf{Fed}erated Learning with \textbf{G}roup \textbf{D}ebiasing), which discovers latent preference groups during federated training and learns a single reward model using group-debiased client sampling. By counteracting the effect of group imbalance, \alg\ learns an initialization that remains highly adaptable, enabling effective personalization without prior knowledge of the underlying groups.

\end{abstract}

\maketitle
\pagestyle{plain}

\section{Introduction}

Large language models (LLMs) are increasingly deployed in human-facing systems, where alignment with human preferences is crucial to user safety and satisfaction~\citep{bai2022constitutional,askell2021general,casper2023open}. A standard alignment pipeline typically trains a \emph{reward model} to predict human preferences and then optimizes the LLM through RLHF methods~\citep{schulman2017proximal,ouyang2022training}. However, preference data are highly sensitive and often cannot be centralized due to data-protection regulations~\citep{regulation2016regulation, illman2019california} and cross-jurisdictional sharing constraints~\citep{kopf2023openassistant}. Federated learning (FL)~\citep{mcmahan2017communication,li2020fedprox,kairouz2021advances}
offers a practical alternative: data remain local while a coordinating server aggregates model updates from decentralized clients.

Existing FL-based preference alignment methods~\citep{fan2024fedrlhf,wu2024towards} struggle with \emph{preference heterogeneity} because they ultimately produce a single global LLM. For instance, FedRLHF~\citep{fan2024fedrlhf} shows clear performance degradation under severe heterogeneity. Even FedBiscuit~\citep{wu2024towards}, which uses multiple server-side reward models to label unlabeled response pairs, aggregates these signals into a single consensus to train one LLM via DPO~\citep{rafailov2023direct}. Consequently, existing methods fail to represent individual user preferences, underscoring the need for \emph{personalized reward models} tailored to each client.

Recent personalized FL (PFL) methods~\citep{oh2021fedbabu,dong2022spherefed,kim2023fedfn,kim2025feddr} widely adopt a two-stage strategy: pre-training a single shared model globally and fine-tuning it locally per client. While effective under standard data heterogeneity, where clients share label consensus despite input distribution shifts, this approach struggles under preference heterogeneity, where different clients may assign opposing labels to the same response pair. This fundamental conflict raises a question about whether a single shared initialization can still provide an effective starting point for personalized reward modeling.

\begin{figure*}[t]
    \centering
    \includegraphics[width=0.88\textwidth]{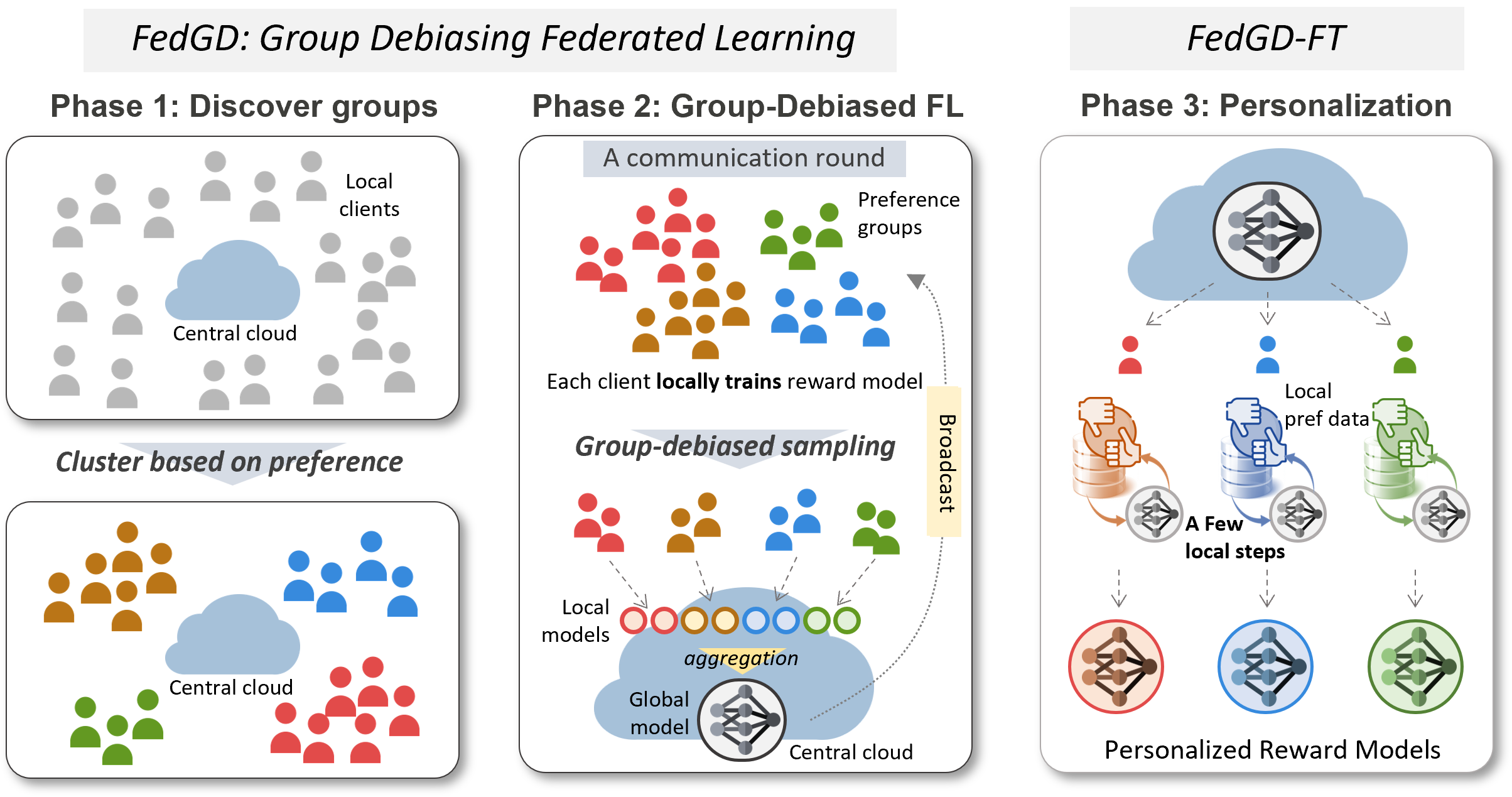}
    \vspace{-9pt}
    \caption{Overview of \alg. \textbf{Phase 1} discovers the preference groups, which can differ in size. \textbf{Phase 2} trains a single reward model over them with group-debiased client sampling, which compensates for group-size imbalance during client selection. The resulting model serves as a shared initialization for personalization. \textbf{Phase 3} personalizes the initial model for each client through a few local optimization steps, yielding \algft.}
    \Description{Three-panel diagram: clients grouped into four preference groups of unequal size, a federated round drawing the same number of clients from every group into one shared model, and that model fine-tuned locally into personalized models.}
    \label{fig:algo}
\end{figure*}

Existing federated methods that explicitly address conflicting preferences or tasks instead avoid this question by clustering similar clients and training one single model per group. For example, FedBiscuit~\citep{wu2024towards} groups clients according to preference similarity, while FedLEASE~\citep{wang2025adaptive} clusters clients across related tasks, so that each client is initialized from a model that never averages over conflicting objectives. However, whether these group-specific initializations actually lead to better personalized reward models than a single shared initialization remains unknown. \textbf{\emph{We therefore ask: what makes a good initialization for personalized reward modeling under preference heterogeneity?}}



We find that a group-specific initialization is not the answer. Under balanced preference groups, a single global model trained with FedAvg~\citep{mcmahan2017communication} performs at the level of random guessing before adaptation, yet within a few local steps for personalization, it quickly surpasses both centralized training (CL) and a separate centralized model per ground-truth preference group (Multi-CL).

Our key hypothesis is that preference learning contains two distinct components. The first is \emph{shared knowledge}: identifying the latent feature that distinguishes the two responses. All clients agree on this distinction, so collaborative FL reinforces a common representation of it. The second is \emph{client-specific knowledge}: deciding which side of the latent feature should be preferred. Since this preference differs across clients, no single decision boundary can be optimal for everyone, explaining the poor pre-adaptation accuracy of the global model trained by FL. Consequently, the goal of federated learning should not be to produce the final personalized model, but rather an initialization that captures the shared knowledge while remaining readily adaptable to each client's preference.

To diagnose such an initialization, we adopt the \emph{Gradient Quotient} ($GQ$)~\citep{dauphin2019metainit}, which measures how rapidly the local gradient changes. A small $GQ$ indicates that local optimization remains stable, allowing successive personalization steps to reinforce one another. Such an initialization is therefore well suited for rapidly adapting to each client's local preferences. In our experiments, FedAvg converges to initializations with substantially lower $GQ$ than CL or Multi-CL, which explains why it serves as a better starting point for personalization despite its low start accuracy.

However, when preference groups are imbalanced, the cancellation becomes asymmetric, causing the shared model to become biased toward the majority preferences and reducing its ability to serve as a good initialization for minority clients. We find that balancing client sampling across preference groups during federated training restores the personalization capability of the resulting initialization. By this key observation, we propose \alg\ (\textbf{Fed}erated Learning with \textbf{G}roup \textbf{D}ebiasing), which first discovers the preference groups during FL and then trains a single reward model under group-debiased sampling over the discovered partition (Figure~\ref{fig:algo}).

\textbf{Our contributions are as follows:} 
\begin{itemize}
    \item We introduce an optimization-based perspective on what makes a good initialization for efficient personalization in reward modeling: an initialization
    adapts well when the loss around it is flat, so that successive local steps reinforce one another. We measure this flatness with $GQ$, which tracks how much the local
    gradient changes after a single step.
    
    \item We construct a synthetic benchmark with configurable preference groups and group-size imbalance, enabling controlled analysis for personalized reward modeling under federated settings.
    
    \item We show that, under balanced preference groups, a single FedAvg model provides a better initialization for personalization than both CL and Multi-CL despite its poor
    pre-adaptation accuracy, and identify group imbalance as the condition under which
    this advantage weakens.
    
    \item We propose \alg, which discovers latent preference groups during federated learning and performs group-debiased client sampling to recover high-quality
    initializations without requiring prior knowledge of the true preference groups.
    
\end{itemize}

\section{Preliminaries}
\label{sec:prelim}
We study federated personalized reward modeling, where clients collaboratively train initial global reward models on decentralized pairwise preferences and then fine-tune them locally. We first formalize this workflow (Section~\ref{subsec:fl-procedure}) and then propose two key metrics to evaluate the personalization potential of initial global models (Section~\ref{sec:metrics}). Finally, Section~\ref{sec:ex_design} details the experimental setup.

\subsection{Federated Reward Modeling and Personalization}
\label{subsec:fl-procedure}
We study FL~\cite{mcmahan2017communication} over $C$ clients with pairwise preference data. Each client $c$ holds a local dataset $D_{\text{train}}^c = \{(x_i^c, y_i^{c,+}, y_i^{c,-})\}_{i=1}^{N_c}$, where $x_i^c$ is a prompt, $y_{i}^{c,+}$ the preferred response, and $y_{i}^{c,-}$ the less-preferred one. Let $r_\theta(x, y)$ denote the reward the model assigns to response $y$ for prompt $x$. For a preference $(x, y^{+}, y^{-})$, the \emph{reward margin} $m_\theta(x, y^{+}, y^{-}) = r_\theta(x, y^{+}) - r_\theta(x, y^{-})$ is the signed gap between the two responses, so the prediction is correct when $m_\theta > 0$ and incorrect when $m_\theta \le 0$. Clients minimize the pairwise preference loss $\ell_\theta = -\log \sigma(m_\theta)$, which drives the margin positive, giving the full-batch training loss $\mathcal{L}_c(\theta) = \frac{1}{N_c}\sum_i \ell_\theta(x_i^c, y_i^{c,+}, y_i^{c,-})$ of client $c$.

FL proceeds over $R$ communication rounds.
At the beginning of round $r$, the server broadcasts the current global parameters $\theta^{(r-1)}$ to a sampled subset of clients $S_r \subset [C]$.
Each selected client $c \in S_r$ performs $\tau$ local training iterations on $D_{\text{train}}^c$ with batch size $B$, and returns updated parameters $\theta_{c}^{(r)}$.
The server aggregates these local updates via weighted averaging~\cite{li2019convergence, wu2024towards} to obtain the new global parameters $\theta^{(r)}$. The description above corresponds to a single-global design in which the server maintains a single parameter vector $\theta^{(r)}$. Some of the designs we compare instead maintain $K$ global models $\{\theta_k^{(r)}\}_{k=1}^K$ on the server and apply the same broadcast--update--aggregate procedure independently to each model over its associated subset of clients.

After $R$ training rounds in the \emph{global} phase, we run a separate \emph{personalization} phase. Each client $c$ receives a single initialization $\theta_c^{\text{init}}$, fine-tunes it on its own preference data $D_{\text{train}}^c$ without further communication, and obtains a personalized reward model $\theta_c^{\text{PFL}}$. Since local adaptation runs on the client, where compute is limited, the number of local steps an initialization requires is itself a cost. We therefore report accuracy after a small number of local steps alongside the accuracy eventually reached. How $\theta_c^{\text{init}}$ is constructed is the design choice we study, and we specify it for each method in Sections~\ref{sec:motivation} and~\ref{sec:exp}. Throughout the paper, we refer to the models that serve as initializations for personalization as \emph{global} models and to $\theta_c^{\text{PFL}}$ as the resulting \emph{personalized} reward model for client $c$.

\subsection{Quantifying Personalization Capability}
\label{sec:metrics}
Given a reward model $\theta$ as the \emph{initialization} for
personalization, we characterize its personalization capability by estimating how effectively a small number of local updates can improve it. For this, we introduce two diagnostics: (i) \emph{Gradient Quotient} ($GQ$)~\cite{dauphin2019metainit}, a model-level measure of \emph{flatness} that assesses whether consecutive local steps reinforce one another, and (ii) \emph{Headroom} ($H$), a per-sample measure that we propose to assess whether a local update moves an individual sample toward its correct preference. Both quantities are computed using each client's local preference data.

\noindent\textbf{Gradient Quotient ($GQ$): flatness of the initialization.}\quad 
A good initialization should be one from which local updates consistently move toward the client's optimum rather than changing gradient dramatically after one update. We measure this property using the $GQ$~\cite{dauphin2019metainit}, which quantifies how much the gradient changes after a single optimization step.

Let
$g^c(\theta) = \nabla \mathcal{L}_c(\theta)$ be the gradient of client $c$'s
full-batch training loss and $g^{c,l}$ its restriction to the LoRA parameters
$\theta^l$ of layer $l$. The layer-wise $GQ$ is
\begin{align*}
GQ(c, l)
&= \frac{1}{\#\theta^l}
\left\lVert \frac{g^{c,l}\!\big(\theta - \eta\, g^c(\theta)\big)}{g^{c,l}(\theta)} - \mathbf{1} \right\rVert_1
\approx
\frac{\eta}{\#\theta^l}
\left\lVert \frac{[\nabla^2\mathcal{L}_c(\theta)\, g^c(\theta)]^{\,l}}{g^{c,l}(\theta)} \right\rVert_1 ,
\end{align*}
where $\#\theta^l$ is the number of LoRA parameters in layer $l$. The first-order
approximation shows that $GQ$ is determined by the Hessian–gradient product, which captures the local curvature along the update direction. Consequently, $GQ$ increases with the curvature encountered by gradient descent. A small $GQ$ indicates that the gradient remains nearly unchanged in both direction and magnitude, allowing successive local updates to remain aligned and accumulate toward the client's optimum. In contrast, a large $GQ$ indicates that the gradient is rapidly distorted after a single step, causing subsequent steps to deviate from the original descent direction and partially cancel earlier progress, thereby reducing the effectiveness of local personalization.

We obtain a layer-wise statistic by averaging over clients,
$GQ(l) = \tfrac{1}{C}\sum_c GQ(c, l)$ over the $24$ LoRA-adapted layers of
Qwen2-0.5B~\cite{yang2024qwen2technicalreport}, our base model
(Section~\ref{sec:exp}). Studies of few-step adaptation commonly separate a
shared representation body from task-specific upper layers, both in
meta-learning~\cite{oh2021boil} and in personalized FL~\cite{oh2021fedbabu}, so we
report a Body average $GQ_{\mathcal{B}}$ over the front $16$ layers and a Head
average $GQ_{\mathcal{H}}$ over the back $8$.

\noindent\textbf{Headroom ($H$): one-step correctability of a sample.}\quad
$GQ$ measures whether local updates are stable, not whether they lead to
the right direction. We define \emph{Headroom} to answer the latter for a single
sample. Starting from the initialization under evaluation,
$\theta_0 = \theta_c^{\text{init}}$, we take one full-batch \emph{training} step on
client $c$'s data, $\theta_1 = \theta_0 - \eta\, g^{c}(\theta_0)$, and define the
Headroom as the first-order effect of this step on the margin of a held-out
\emph{test} sample $(x, y^{+}, y^{-})$ of the same client:
\begin{equation*}
H(x, y^{+}, y^{-})
= -\eta\, \big\langle \nabla_\theta m_\theta\big|_{\theta_0},\ g^{c}(\theta_0) \big\rangle
\approx \Delta m,
\end{equation*}
where $\Delta m = m_{\theta_1} - m_{\theta_0}$ is the true margin change. A
positive $H$ means one training step moves the held-out sample toward its correct
preference, and a larger $H$ indicates more room to move it.

\begin{table*}[t]
\centering
\small
\setlength{\tabcolsep}{3.5pt}
\renewcommand{\arraystretch}{0.95}
\caption{Diagnostics and personalized accuracy under balanced settings. $GQ(l)$: mean$\pm$std over the 24 layers, with Body/Head averages $GQ_{\mathcal{B}}, GQ_{\mathcal{H}}$. $\mathcal{R}$ / $\mathcal{W}$: test samples initially correct / wrong at $\theta_0$. $m^{0}$: mean reward margin at $\theta_0$. $H^{+}$: fraction with positive Headroom. $P^{10}$: fraction correct after 10 local steps. $\mathrm{acc}^{0}$ in parentheses. Highlighting the best $\mathrm{acc}^{10}$ and bold the best per column.}
\label{tab:balanced}
\vspace{-8pt}
\begin{tabular}{lccc cc cc cc c}
\toprule
& \multicolumn{3}{c}{Flatness ($\downarrow$)}
& \multicolumn{2}{c}{Margin ($m^{0}$)}
& \multicolumn{2}{c}{Retention ($\mathcal{R}$, \%, $\uparrow$)}
& \multicolumn{2}{c}{Correction ($\mathcal{W}$, \%, $\uparrow$)}
& \multicolumn{1}{c}{Accuracy (\%, $\uparrow$)} \\
\cmidrule(lr){2-4} \cmidrule(lr){5-6} \cmidrule(lr){7-8} \cmidrule(lr){9-10} \cmidrule(lr){11-11}
method & $GQ(l)$ & $GQ_{\mathcal{B}}$ & $GQ_{\mathcal{H}}$
       & $m^{0}_{\mathcal{R}}$ & $m^{0}_{\mathcal{W}}$
       & $H^{+}_{\mathcal{R}}$ & $P^{10}_{\mathcal{R}}$
       & $H^{+}_{\mathcal{W}}$ & $P^{10}_{\mathcal{W}}$
       & $\mathrm{acc}^{10}\,(\mathrm{acc}^{0})$ \\
\midrule
CL       & $1.14 \pm 0.24$ & 1.29 & 0.84
         & $+16.00$ & $-15.77$
         & 34.18 & 82.32 & 62.02 & 21.49
         & 52.15 (50.45) \\
Multi CL & $0.86 \pm 0.23$ & 1.00 & 0.56
         & $+22.88$ & $-12.39$
         & 13.01 & \textbf{97.28} & 59.51 & 25.77
         & 91.45 (92.20) \\
\rowcolor{flhl}
FL       & $\mathbf{0.10 \pm 0.01}$ & \textbf{0.11} & \textbf{0.09}
         & $+2.38$ & $-2.34$
         & \textbf{56.94} & 94.01 & \textbf{96.00} & \textbf{91.99}
         & \textbf{93.45 (50.40)} \\
\bottomrule
\end{tabular}
\end{table*}

We compute $H$ for every sample in the client's held-out test split. Since
correctability depends on whether a sample is already correct, we partition the
test set at $\theta_0$ into the \emph{retention} set $\mathcal{R} = \{m_{\theta_0} > 0\}$
and the \emph{correction} set $\mathcal{W} = \{m_{\theta_0} \le 0\}$, and report for
each set the mean Headroom $H^{0}$ and the fraction of samples with positive
Headroom $H^{+}$, averaged over clients.

\subsection{Datasets and Preference Groups}
\label{sec:ex_design}
We evaluate on two datasets: a real-world set with natural annotator preferences, and a synthetic set with controlled style heterogeneity. For each client, we reserve 50 samples each for validation and testing, and use the rest for training.

\noindent\textbf{Real-world dataset.}\quad
We use the Reddit TL;DR summarization dataset~\citep{stiennon2020learning, volske2017tl}, which contains human preference annotations along with client IDs. We filter 144,502 samples for 34 clients who participate in all three train/valid1/valid2 splits with at least 100 samples each. Each client corresponds to an actual human annotator, and the dataset inherently exhibits natural data imbalance and genuine preference heterogeneity, with the underlying preference groups being \emph{unknown}.

\noindent\textbf{Synthetic dataset.}\quad
We gather prompts by combining UltraFeedback~\citep{cui2023ultrafeedback} and p-Soups~\citep{jang2023personalized}, resulting in 52K unique prompts. Using GPT-4o-mini,\footnote{\url{https://platform.openai.com/docs/models/gpt-4o-mini}}we generate two candidate responses per prompt along two style axes---(1) \emph{Elementary} vs.\ \emph{PhD-level} and (2) \emph{Humorous} vs.\ \emph{Non-humorous}---allocating 26K prompts per axis. We then simulate 40 clients (650 samples per axis), grouped into the four preference groups formed by the two axes: (G1) Elementary/Humorous, (G2) Elementary/Non-humorous, (G3) PhD/Humorous, and (G4) PhD/Non-humorous. Clients are indexed by group, starting with G1 and progressing through G4 in sequence.

\noindent\textbf{Group-size balance.}\quad
Using the synthetic dataset, we evaluate how effectively the initial shared model adapts to individual clients under varying group-size distributions. We consider two group-size configurations across G1, G2, G3, and G4:
\begin{itemize}\setlength\itemsep{1pt}
    \item \textbf{Balanced (10/10/10/10):} 10 clients per group, maintaining an equal balance across both preference axes.
    \item \textbf{Imbalanced (15/15/5/5):} the Elementary-dominant groups (G1, G2) are overrepresented while the PhD-dominant groups (G3, G4) are underrepresented, naturally biasing the shared model toward the majority preference.
\end{itemize}

\section{A Single Federated Model Suffices, Until Group Imbalance}
\label{sec:motivation}

In this section, we compare three initializations in terms of the personalization capability of the resulting models. Interestingly, our finding indicates that FL provides a stronger initialization when preference groups are well balanced. Results show that this advantage arises because the local training loss is relatively flat around the initialization, while most test samples retain positive Headroom toward the correct preference. However, this benefit weakens under group imbalance, where FL becomes increasingly biased toward majority preferences and adaptation slows sharply. These findings motivate \alg, our proposed approach.

\noindent\textbf{Compared initializations.}\quad
We compare 3 ways of producing the reward model that each client fine-tunes locally, trained for 200 communication rounds with the same number of updates.

\begin{itemize}
\setlength\itemsep{2pt}
    \item \textbf{CL:} pools all preference data on one server and trains a single centralized model, so conflicting preferences are mixed \emph{within every batch}.
    \item \textbf{FL:} trains a single global model with FedAvg~\cite{mcmahan2017communication}, so the conflict is concentrated
    \emph{at aggregation}.
    \item \textbf{Multi-CL:} trains one centralized model per ground-truth preference group, so \emph{no conflict arises within a model}. Each client is initialized from its own group's model. Since the true groups are unknown in practice, Multi-CL is a reference point rather than a deployable method.
\end{itemize}

\begin{figure}
    \centering
    \includegraphics[width=0.8\linewidth]{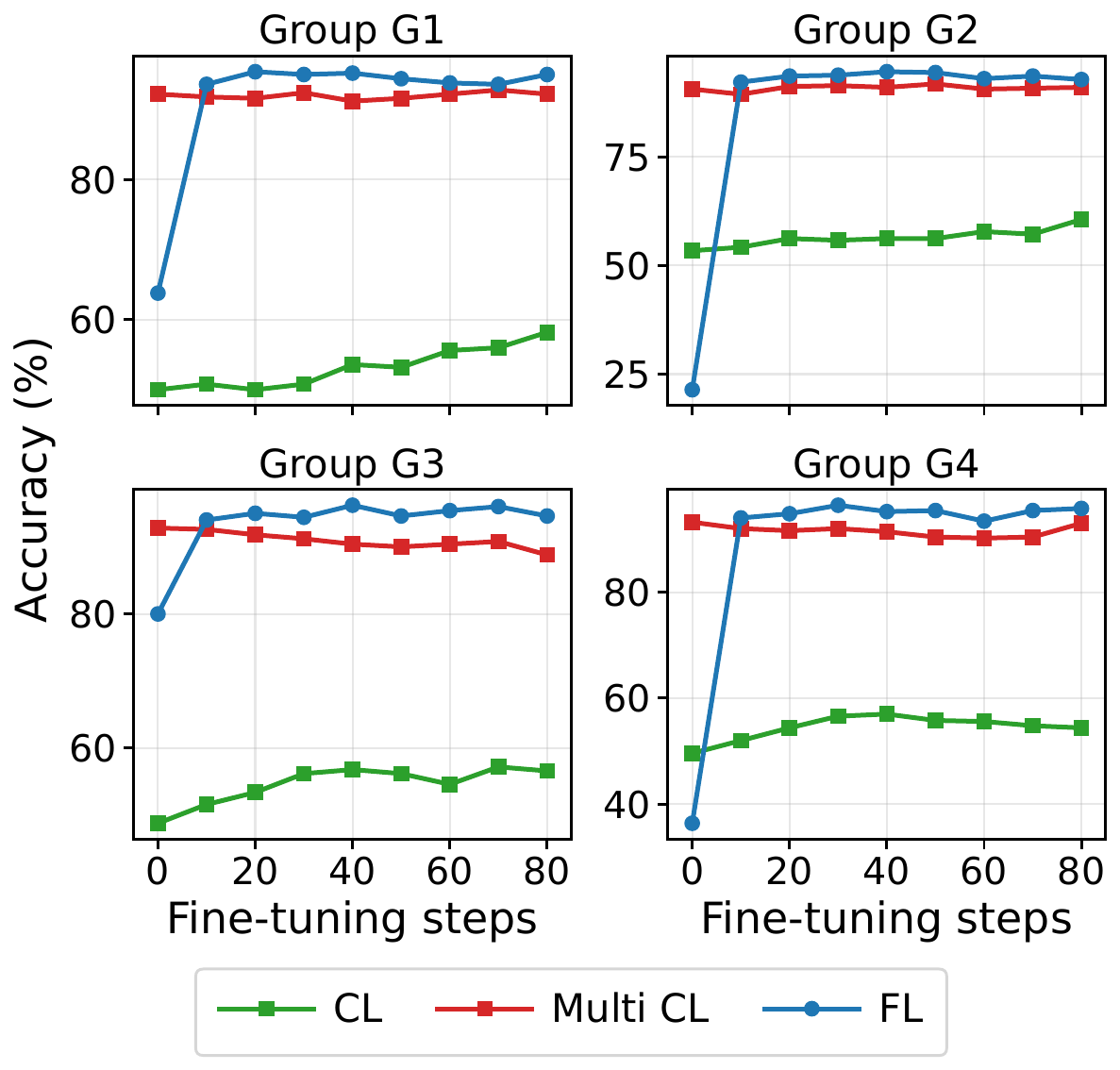}
    \vspace{-12pt}
    \caption{Group-wise personalized accuracy under the balanced setting, evaluated every 10 steps up to 80 local fine-tuning steps.}
    \Description{}
    \label{fig:groupwise_balance}
\end{figure}

\begin{table*}[t]
\centering
\small
\setlength{\tabcolsep}{3.5pt}
\renewcommand{\arraystretch}{0.95}
\caption{Diagnostics and personalized accuracy under the imbalanced setting
(15/15/5/5); the first row repeats balanced FL for reference.
$GQ(l)$: mean$\pm$std over the 24 LoRA layers, with Body/Head averages
$GQ_{\mathcal{B}}, GQ_{\mathcal{H}}$. $\mathcal{R}$ / $\mathcal{W}$: test
samples initially correct / wrong at $\theta_0$. $m^{0}$: mean reward margin
at $\theta_0$. $H^{+}$: fraction with positive headroom. $P^{10}$: fraction
correct after 10 local steps. Highlighting marks the best
$\mathrm{acc}^{10}$; bold the best per column among the imbalanced rows.}
\label{tab:imbalanced}
\vspace{-8pt}
\begin{tabular}{lccc cc cc cc c}
\toprule
& \multicolumn{3}{c}{Flatness ($\downarrow$)}
& \multicolumn{2}{c}{Margin ($m^{0}$)}
& \multicolumn{2}{c}{Retention ($\mathcal{R}$, \%, $\uparrow$)}
& \multicolumn{2}{c}{Correction ($\mathcal{W}$, \%, $\uparrow$)}
& \multicolumn{1}{c}{Accuracy (\%, $\uparrow$)} \\
\cmidrule(lr){2-4} \cmidrule(lr){5-6} \cmidrule(lr){7-8} \cmidrule(lr){9-10} \cmidrule(lr){11-11}
method & $GQ(l)$ & $GQ_{\mathcal{B}}$ & $GQ_{\mathcal{H}}$
       & $m^{0}_{\mathcal{R}}$ & $m^{0}_{\mathcal{W}}$
       & $H^{+}_{\mathcal{R}}$ & $P^{10}_{\mathcal{R}}$
       & $H^{+}_{\mathcal{W}}$ & $P^{10}_{\mathcal{W}}$
       & $\mathrm{acc}^{10}\,(\mathrm{acc}^{0})$ \\
\midrule
\rowcolor{flhl}
FL (balanced) & $0.10 \pm 0.01$ & 0.11 & 0.09
              & $+2.38$ & $-2.34$
              & 56.94 & 94.01 & 96.00 & 91.99
              & 93.45 (50.40) \\
\midrule
CL         & $1.09 \pm 0.28$ & 1.25 & 0.76
           & $+20.41$ & $-19.42$
           & 25.72 & 86.50 & 69.61 & 20.14
           & 60.50 (59.85) \\
Multi CL   & $0.95 \pm 0.25$ & 1.11 & 0.63
           & $+23.96$ & $-11.79$
           & 10.44 & 97.60 & 58.42 & 26.14
           & 91.50 (92.10) \\
FL         & $0.72 \pm 0.25$ & 0.88 & 0.39
           & $+27.90$ & $-27.02$
           & 3.08 & \textbf{99.11} & \textbf{96.74} & 2.09
           & 61.75 (61.80) \\
\rowcolor{fthl}
FL\_target & $\mathbf{0.27 \pm 0.02}$ & \textbf{0.28} & \textbf{0.24}
           & $+3.59$ & $-3.65$
           & \textbf{75.78} & 93.42 & 93.44 & \textbf{90.62}
           & \textbf{92.35 (54.95)} \\
\bottomrule
\end{tabular}
\end{table*}

\subsection{FL Suffices Under Balanced Groups}
\label{sec:motivation-balanced}
Under the balanced setting, FL shows the best initialization for personalization. Let $\mathrm{acc}^{0}$ denote the accuracy of an initialization before any local update and $\mathrm{acc}^{10}$ the accuracy after 10 local fine-tuning steps, both averaged over clients. Before fine-tuning, Multi-CL reaches $\mathrm{acc}^{0} = 92.20$, while CL and FL reach only $50.45$ and $50.40$, the accuracy of random guessing on binary preference pairs (Table~\ref{tab:balanced}). Figure~\ref{fig:groupwise_balance} shows that this average hides a wide spread for FL, whose initial accuracy ranges from about $20\%$ on G2 to about $80\%$ on G3. However, just 10 local steps change the picture: FL reaches $\mathrm{acc}^{10} = 93.45$ and overtakes Multi-CL, which ends at $91.45$, slightly below where it started, while CL improves only to $52.15$. The same pattern appears in every group: FL gains almost all of its improvement within 10 steps, whereas CL improves only gradually over 80 steps and Multi-CL changes little or slightly declines.


FL gains the most from a few local steps despite its low initial accuracy, and
Table~\ref{tab:balanced} attributes this to a loss that is flat across all layers
around the FL model. The flatness follows from how far each initialization has
already committed. Every client must detect what distinguishes the two responses---here the style axis---to fit its own labels, and what differs is only which side it prefers, so averaging the locally trained models reinforces the shared detection while the opposing side choices offset one another. The resulting model therefore sits close to the decision boundary, with initial margins of only $m^{0}_{\mathcal{R}} = +2.38$ and
$m^{0}_{\mathcal{W}} = -2.34$, where the loss $-\log\sigma(m)$ is far from
saturation and its gradient changes slowly. CL cannot separate detecting the axis
from choosing a side, since conflicting preferences enter the same batch, and
Multi-CL removes the conflict entirely by training each model on a quarter of the
clients that share one direction; both commit to a side and reach
$m^{0}_{\mathcal{R}} = +16.00$ and $+22.88$, deep in the saturated regime where a
single step distorts the gradient. This is reflected in the diagnostics: FL is an
order of magnitude flatter than both baselines at every layer
($GQ_{\mathcal{B}} = 0.11$ vs.\ $1.29$ and $1.00$; $GQ_{\mathcal{H}} = 0.09$
vs.\ $0.84$ and $0.56$), and the gap is largest in the Body, the layers that
detect the axis.

On the flat FL initialization, fine-tuning corrects wrong predictions without
sacrificing the correct ones. For FL, $96.00\%$ of the wrong set $\mathcal{W}$
carries positive headroom ($H^{+}_{\mathcal{W}}$), and after 10 steps $91.99\%$ of
$\mathcal{W}$ is answered correctly ($P^{10}_{\mathcal{W}}$), while $94.01\%$ of
the right set $\mathcal{R}$ stays correct ($P^{10}_{\mathcal{R}}$). CL and Multi-CL
start with less headroom on $\mathcal{W}$ ($62.02\%$ and $59.51\%$) and correct
only $21.49\%$ and $25.77\%$ of it, even though Multi-CL keeps slightly more of
$\mathcal{R}$ than FL ($97.28\%$).

\vspace{-10pt}
\subsection{Group-Debiased Sampling Restores FL Under Imbalance}
\label{sec:motivation-imbalanced}
\begin{figure}[t]
    \centering
    \includegraphics[width=0.8\linewidth]{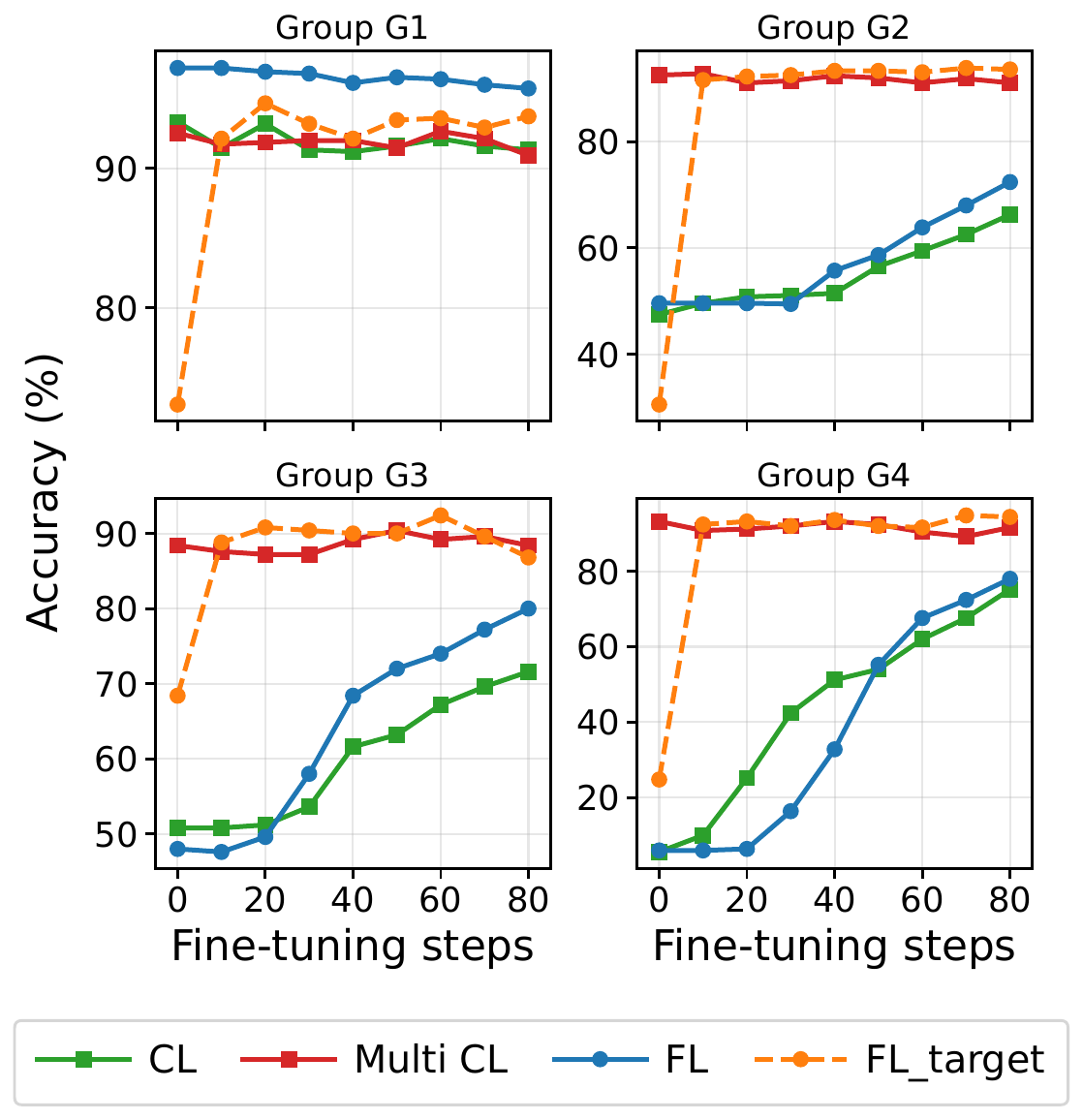}
    \vspace{-8pt}
    \caption{Group-wise PFL accuracy under the imbalanced setting for the four methods (CL, Multi CL, FL, and FL\_target), evaluated every 10 steps up to 80 fine-tuning steps.}
    \label{fig:groupwise_imbalance}
\end{figure}

Under group imbalance, FL adapts more slowly. Although FL begins with a substantially higher accuracy than in the balanced case, 10 local steps leave it essentially unchanged. FL starts in $\mathrm{acc}^{0} = 61.80$, compared with $50.40$ under balanced groups, yet reaches only $\mathrm{acc}^{10} = 61.75$~(Table~\ref{tab:imbalanced}). Figure~\ref{fig:groupwise_imbalance} shows that the initial accuracy again varies widely across groups, from near $10\%$ on G4 to $98\%$ on G1, but the gains from fine-tuning are smaller than under balanced groups, where every group rose above $90\%$ within 10 steps. G2, G3, and G4 instead climb slowly and stay below G1 even after 80 steps.

To test the hypothesis that the degradation in personalization is caused by group imbalance rather than the FL procedure itself, we consider a group-debiased sampling strategy that prevents any preference group from being underrepresented in a communication round. Specifically, we modify FL only in the client sampling step while leaving the rest of the training procedure unchanged, denoted by FL\_target. In our experiments, each round samples $|S_r| = 5$ clients: one from each of the four groups and one additional client chosen uniformly at random from all groups. This simple modification lifts accuracy from $\mathrm{acc}^{0} = 54.95$ to $\mathrm{acc}^{10} = 92.35$, a level standard FL does not reach within the same number of local steps ($\mathrm{acc}^{0} = 61.80$ to $\mathrm{acc}^{10} = 61.75$) (Table~\ref{tab:imbalanced}). Figure~\ref{fig:groupwise_imbalance} shows FL\_target exceeding $90\%$ on every group within 10 local steps, a behavior FL showed only in the balanced setting. On this dataset, \alg\ with $K=4$ recovers the four ground-truth groups exactly, so FL\_target coincides with \alg\ (Section~\ref{sec:method}). 

Group-debiased sampling works by keeping the loss uniformly flat across layers,
unlike uniform sampling. Under FL, the Body degrades far more than the Head, with
$GQ_{\mathcal{B}}$ rising from $0.11$ in the balanced setting to $0.88$ and
$GQ_{\mathcal{H}}$ from $0.09$ to $0.39$ (Table~\ref{tab:imbalanced}). Correction
suffers as well: $96.74\%$ of the wrong set $\mathcal{W}$ still carries positive
headroom, yet only $2.09\%$ of $\mathcal{W}$ is answered correctly after 10 steps,
against $91.99\%$ in the balanced setting. The direction of the update is therefore right but its reach is not: imbalanced FL
commits to the majority side and starts at $m^{0}_{\mathcal{W}} = -27.02$, an
order of magnitude beyond the balanced case ($-2.34$), so ten steps cannot close
the gap. The right set $\mathcal{R}$ changes little instead, with $99.11\%$ staying correct but only $3.08\%$ carrying positive headroom. FL\_target starts lower than FL at $\mathrm{acc}^{0} = 54.95$, yet keeps
the Body and the Head close ($GQ_{\mathcal{B}} = 0.28$, $GQ_{\mathcal{H}} = 0.24$) and stays near the decision
boundary ($m^{0}_{\mathcal{W}} = -3.65$ against $-27.02$ for FL), raises $H^{+}_{\mathcal{R}}$ from $3.08\%$ to $75.78\%$, and recovers correction to
$P^{10}_{\mathcal{W}} = 90.62\%$. Group-debiased sampling is therefore what lets FL adapt within a few local steps
under group imbalance. However, FL\_target assumes access to the true preference groups, which are unknown in practice.
This naturally raises the following question: \emph{``Can group debiasing be achieved without knowing the true preference groups?''}
\section{Federated Learning with Group Debiasing}
\label{sec:method}
In this section, we propose \alg, which applies group debiasing when the true groups
are unknown. \alg proceeds in two phases: (1) discovering the groups during
federated learning, and (2) training a single reward model with group-debiased
sampling over the discovered groups. Each client then fine-tunes the resulting
model following the personalization phase of Section~\ref{subsec:fl-procedure},
yielding the personalized reward models we denote by \algft. The full algorithm is
given in Appendix~\ref{app:fedgd}.

\begin{table*}[t]
\centering
\small
\setlength{\tabcolsep}{4.5pt}
\renewcommand{\arraystretch}{0.95}
\caption{PFL accuracy comparison over clients from a single seed.
$\mathrm{acc}^{t}$ is the accuracy after $t$ local fine-tuning steps, reported
as mean$_{\pm\text{std}}$; $\mathrm{acc}^{0}$ is given as the mean only.
Multi CL requires known groups and is not deployable on real annotators
(``--''). \textbf{Bold} marks the best value in each post-adaptation column.}
\label{tab:pfl}
\vspace{-8pt}
\resizebox{0.9\textwidth}{!}{%
\begin{tabular}{l ccc ccc ccc}
\toprule
& \multicolumn{3}{c}{Balanced (Synthetic)}
& \multicolumn{3}{c}{Imbalanced (Synthetic)}
& \multicolumn{3}{c}{Real-world} \\
\cmidrule(lr){2-4} \cmidrule(lr){5-7} \cmidrule(lr){8-10}
Algorithm & $\mathrm{acc}^{0}$ & $\mathrm{acc}^{10}$ & $\mathrm{acc}^{80}$
          & $\mathrm{acc}^{0}$ & $\mathrm{acc}^{10}$ & $\mathrm{acc}^{80}$
          & $\mathrm{acc}^{0}$ & $\mathrm{acc}^{80}$ & $\mathrm{acc}^{240}$ \\
\midrule
Local
  & 49.20 & $\acc{49.95}{7.26}$ & $\acc{58.05}{7.69}$
  & 50.25 & $\acc{51.30}{7.44}$ & $\acc{58.70}{7.07}$
  & 51.24 & $\acc{52.12}{6.65}$ & $\acc{54.59}{7.66}$ \\
\midrule
CL
  & 51.80 & $\acc{53.70}{7.15}$ & $\acc{55.40}{5.99}$
  & 56.00 & $\acc{56.90}{10.89}$ & $\acc{62.60}{7.93}$
  & 58.06 & $\acc{57.06}{9.08}$ & $\acc{61.00}{6.69}$ \\
Multi CL
  & 92.10 & $\acc{92.15}{3.88}$ & $\acc{91.60}{4.29}$
  & 92.30 & $\acc{91.70}{3.63}$ & $\acc{91.60}{3.67}$
  & -- & -- & -- \\
FL~\cite{mcmahan2017communication}
  & 50.35 & $\acc{92.45}{3.76}$ & $\mathbf{\acc{94.70}{2.92}}$
  & 55.60 & $\acc{90.55}{5.20}$ & $\mathbf{\acc{93.80}{3.40}}$
  & 60.82 & $\acc{61.53}{7.70}$ & $\acc{63.00}{9.10}$ \\
\midrule
FedBiscuit~\cite{wu2024towards}
  & 68.80 & $\acc{66.90}{6.18}$ & $\acc{71.45}{10.35}$
  & 78.45 & $\acc{77.85}{9.85}$ & $\acc{80.80}{8.46}$
  & 59.12 & $\acc{58.53}{6.76}$ & $\acc{57.06}{5.77}$ \\
Soft-FL
  & 51.70 & $\acc{62.40}{7.51}$ & $\acc{71.30}{6.09}$
  & 56.50 & $\acc{75.95}{9.81}$ & $\acc{86.20}{7.41}$
  & 60.88 & $\acc{58.71}{7.45}$ & $\acc{57.00}{9.07}$ \\
\alg (ours)
  & 49.15 & $\mathbf{\acc{94.55}{3.58}}$ & $\acc{94.30}{3.86}$
  & 54.95 & $\mathbf{\acc{92.35}{4.02}}$ & $\acc{93.75}{3.89}$
  & 60.59 & $\mathbf{\acc{61.71}{5.98}}$ & $\mathbf{\acc{64.41}{6.79}}$ \\
\bottomrule
\end{tabular}
}
\end{table*}

\subsection{Phase 1: Discovering the Groups}
We discover the groups with clustered FL~\cite{ghosh2020efficient, sattler2020clustered},
which partitions clients by training several models and letting each client join the
one that fits its data best. Over the first $R/2$ rounds, the server maintains $K$
expert models $\{\theta_k\}_{k=1}^{K}$ together with a reference model $\theta_g$,
and each client is assigned to one expert, forming the groups
$\{\mathcal{A}_k\}_{k=1}^{K}$. Every $T$ rounds the clients are reassigned, and in
the rounds between two reassignments the experts are trained on the clients
currently assigned to them. After round $R/2$ the clients are reassigned once more,
and the resulting groups $\{\mathcal{A}_k^{\star}\}_{k=1}^{K}$ are frozen; empty
clusters are discarded, so $K$ denotes the number of discovered groups. Neither the
experts nor $\theta_g$ is used as an initialization for personalization; they serve
only to discover the groups.

\noindent\textbf{Reassigning clients ($\pi(c)$).}\quad
Clients start from a uniform random assignment. Every $T$ rounds each client
evaluates all $K$ experts on its own validation split and is reassigned to the
expert with the lowest validation loss~\cite{wu2024towards},
\begin{equation*}
\pi(c) \leftarrow \arg\min_{k} \operatorname{ValLoss}(\theta_k, D^c_{\text{val}}),
\qquad
\mathcal{A}_k = \{c : \pi(c) = k\},
\end{equation*}
where $\pi(c)$ denotes the expert assigned to client $c$. Each reassignment also
resets every expert to the current reference model, $\theta_k \leftarrow \theta_g$,
so that the experts do not inherit what they learned from the clients they had
before.

\noindent\textbf{Updating the reference model ($\theta_g$).}\quad
Each round draws clients with group-debiased sampling over the current groups
$\{\mathcal{A}_k\}_{k=1}^{K}$. Specifically, clients are sampled from each group with probability inversely proportional to the group's cardinality, so that the resulting client set has a uniform group composition regardless of group size. We denote the resulting participating set $G_r$, the
\textit{group-debiased} counterpart of the sampled set $S_r$ of
Section~\ref{subsec:fl-procedure}. The reference model averages the returned models
uniformly,
\begin{equation*}
\theta_g^{(r)} = \frac{1}{|G_r|} \sum_{c \in G_r} \theta^{(r)}_{\pi(c), c}.
\end{equation*}
Here $\theta_g$ serves as a neutral point to which the experts are reset rather
than as a loss minimizer, and every client takes the same $\tau$ local steps, so
weighting by dataset size would bias the discovered partition toward data volume
rather than preference.

\noindent\textbf{Training the experts ($\theta_k$).}\quad
Every selected client $c \in G_r$ receives the expert $\theta_{\pi(c)}^{(r-1)}$ of its
own group, trains it on its local data, and returns the updated model
$\theta^{(r)}_{\pi(c), c}$ to the server. To estimate each expert from more than the
clients of a single round, the server accumulates the returned models and takes
their cumulative average,
\begin{equation*}
s_k = \frac{1}{n_k} \sum_{r'=r_0}^{r} \sum_{c \in G_{r'} \cap \mathcal{A}_k}
\theta^{(r')}_{k, c},
\qquad
n_k = \sum_{r'=r_0}^{r} |G_{r'} \cap \mathcal{A}_k|,
\end{equation*}
where $r_0$ is the round of the last reassignment. The expert is then updated as a
moving average of its previous value and $s_k$,
\begin{equation*}
\theta_k^{(r)} = (1-w)\, \theta_k^{(r-1)} + w\, s_k .
\end{equation*}


\subsection{Phase 2: Training the Reward Model}
The remaining $R/2$ rounds train a single reward model $\phi$ over the frozen
groups $\{\mathcal{A}_k^{\star}\}_{k=1}^{K}$, starting from a randomly
initialized $\phi^{(R/2)}$. Each round draws clients with group-debiased sampling
and each selected client trains $\phi^{(r)}$ on its local data, returning
$\phi^{(r+1)}_c$. Write $G^{\star}_{r,k}$ for the clients of group $k$ that
participate in round $r$.

Aggregation is \emph{hierarchical}: size-weighted within a group, uniform
across groups:
\begin{equation*}
\bar\phi^{(r+1)}_k
= \sum_{c \in G^{\star}_{r,k}}
  \frac{N_c}{\sum_{c' \in G^{\star}_{r,k}} N_{c'}}\, \phi^{(r+1)}_c ,
\qquad
\phi^{(r+1)}
= \frac{1}{|\mathcal{K}_r|} \sum_{k \in \mathcal{K}_r} \bar\phi^{(r+1)}_k ,
\end{equation*}
where $\mathcal{K}_r$ indexes the groups reached in round $r$. Each group
contributes one model regardless of its size, so the debiasing applied at
sampling is preserved at aggregation.

The resulting $\phi^{(R)}$ is the initial global model from which every client
personalizes using its own local preference data.

\section{Experiments and Results}
\label{sec:exp}
We first evaluate the overall personalization performance of \alg\ on synthetic and real-world datasets. After we verify the proposed mechanism through the diagnostic metrics and examine the robustness of \alg\ with respect to the number of clusters $K$. Additional ablation studies and optimization sensitivity analyses are provided in Appendix~\ref{appsec:additional_results}.

\noindent\textbf{Implementation Details.}\quad Unless explicitly mentioned, all experiments follow the configuration. We run 400 communication rounds of FL, sampling five clients per round ($|S_r| = 5$, $|G_r| = 5$ for our methods), and each selected client performs exactly 30 local training iterations per round with a batch size of 16, using AdamW~\cite{loshchilov2018decoupled} with $(\beta_1, \beta_2) = (0.9, 0.95)$ and a constant learning rate of $1\times10^{-5}$. In configurations with multiple global models and dynamic client--model assignment, the assignments are refreshed every $T=20$ rounds, and FedGD uses a mixing coefficient of $w=0.6$. Methods that maintain multiple server-side models use $K=4$ unless stated otherwise. Qwen2-0.5B~\cite{yang2024qwen2technicalreport} is used as the base model, and LoRA-based parameter-efficient tuning~\cite{hu2022lora, houlsby2019parameter} is applied with rank $r=8$, scaling factor $\alpha=16$, and dropout rate $0.05$. Following the standard Transformer architecture~\cite{vaswani2017attention}, LoRA is injected into the attention projection layers (\texttt{q\_proj}, \texttt{k\_proj}, \texttt{v\_proj}, \texttt{o\_proj}) and the MLP projection layers (\texttt{gate\_proj}, \texttt{up\_proj}) of all 24 Transformer blocks, yielding 24 LoRA-adapted layers over which we compute our layer-wise diagnostics. All experiments are implemented on top of the FedBiscuit~\cite{wu2024towards} codebase.

\noindent\textbf{Baselines.}\quad
Beyond CL, FL, and Multi-CL of Section~\ref{sec:motivation}, we compare three
deployable methods that do not assume known groups.
\begin{itemize}
\setlength\itemsep{2pt}
\item \textbf{Local:} fine-tunes the frozen backbone with LoRA on each client,
without a global phase.
\item \textbf{FedBiscuit}~\citep{wu2024towards}\textbf{:} maintains $K$ expert models on the server. At each communication round, every client receives its assigned expert model, performs local training, and uploads the updated model only to that expert for expert-specific aggregation. Every $T$ communication rounds, clients are reassigned to the expert with the lowest validation loss.
\item \textbf{Soft-FL:} replaces the hard client-to-expert assignment in FedBiscuit with validation-based soft weights over the $K$ experts. At each communication round, every client receives a weighted fusion of the $K$ expert models for local training, and the resulting local model contributes to every expert-specific aggregation according to the client's validation-based weights. See Appendix~\ref{appsec:alg_details} for details. 
\end{itemize}


\subsection{Main Results}

Table~\ref{tab:pfl} compares the deployable methods after 400 communication rounds. \alg attains the highest post-adaptation accuracy in every setting: $\mathrm{acc}^{10}=94.55$ under balanced groups, $92.35$ under imbalance, and $\mathrm{acc}^{240}=64.41$ in the real-world setting. \alg does so from the weakest starting point among all federated methods---$49.15$ and $54.95$ on the synthetic datasets, close to random guessing. The federated phase therefore learns an initialization optimized for rapid personalization rather than high pre-adaptation accuracy.

Group-specific initializations do not necessarily improve personalization and can even degrade real-world performance. Even with ground-truth preference groups, Multi-CL gains only marginally ($92.10\% \to 92.15\%$ at step 10) and eventually degrades ($91.60\%$ after 80 steps). Similarly, in the balanced synthetic setting, FedBiscuit achieves an accuracy of only $66.90\%$ after 10 adaptation steps, far behind standard FedAvg ($92.45\%$). On the real-world dataset, both FedBiscuit ($59.12\% \to 57.06\%$) and Soft-FL ($60.88\% \to 57.00\%$) experience performance degradation during personalization, ultimately finishing below single-model CL ($61.00\%$). These results demonstrate that learning separate initializations is ineffective when preference groups are unknown or imperfectly defined.

\begin{table}[t!]
\centering
\small
\renewcommand{\arraystretch}{1.2}
\caption{Real-world results across different models. $\mathrm{acc}^{240}$ is
reported as mean$_{\pm\text{std}}$ over clients, marked with \textbf{Bold} for the best one, and $\mathrm{acc}^{0}$ as the mean before personalization.}
\label{tab:realworld}
\vspace{-8pt}
\begin{tabular*}{\columnwidth}{@{}l@{\extracolsep{\fill}}cccc@{}}
\toprule
& \multicolumn{2}{c}{Qwen-1.5B}
& \multicolumn{2}{c}{Gemma-2B} \\
\cmidrule(lr){2-3}
\cmidrule(lr){4-5}
Method
& $\mathrm{acc}^{0}$
& $\mathrm{acc}^{240}$
& $\mathrm{acc}^{0}$
& $\mathrm{acc}^{240}$ \\
\midrule
Local
& 49.59
& $\acc{61.12}{8.25}$
& 58.53
& $\acc{67.71}{7.69}$ \\

CL
& 63.71
& $\acc{64.53}{7.44}$
& 64.65
& $\acc{66.41}{7.14}$ \\

FL
& 67.88
& $\acc{68.53}{7.86}$
& 71.29
& $\acc{71.47}{6.98}$ \\

\alg\ (ours)
& 68.65
& $\mathbf{\acc{70.29}{8.35}}$
& 70.29
& $\mathbf{\acc{72.35}{7.44}}$ \\
\bottomrule
\end{tabular*}
\end{table}

\begin{table*}[t]
\centering
\caption{Effect of the number of clusters $K$ on \alg\ under the imbalanced
setting. $GQ(l)$ is the mean$\pm$std over the 24 LoRA layers. $m^{0}$ denotes
the mean reward margin at $\theta_{0}$. $P^{10}$ is the fraction of samples
correctly classified after 10 local steps. Accuracy is reported with the
initial accuracy in parentheses.}
\label{tab:kdiag}
\setlength{\tabcolsep}{3pt}
\resizebox{0.85\textwidth}{!}{%
\begin{tabular}{l ccc cc c c c}
\toprule
& \multicolumn{3}{c}{Flatness ($\downarrow$)}
& \multicolumn{2}{c}{Init.\ margin ($\theta_0$)}
& \multicolumn{1}{c}{Retention (\%, $\uparrow$)}
& \multicolumn{1}{c}{Correction (\%, $\uparrow$)}
& Accuracy (\%, $\uparrow$) \\
\cmidrule(lr){2-4}
\cmidrule(lr){5-6}
\cmidrule(lr){7-7}
\cmidrule(lr){8-8}
\cmidrule(lr){9-9}
Method
& $GQ(l)$ & $GQ_{\mathcal{B}}$ & $GQ_{\mathcal{H}}$
& $m_{\mathcal{R}}^{0}$ & $m_{\mathcal{W}}^{0}$
& $P_{\mathcal{R}}^{10}$
& $P_{\mathcal{W}}^{10}$
& $\mathrm{acc}^{10}\,(\mathrm{acc}^{0})$ \\
\midrule
\alg\ ($K=2$)
& $0.18 \pm 0.02$ & 0.19 & 0.16
& $+3.76$ & $-2.94$
& 94.70
& 85.42
& 91.00 (60.90) \\

\alg\ ($K=3$)
& $0.29 \pm 0.03$ & 0.31 & 0.26
& $+4.23$ & $-4.18$
& 95.20
& 89.72
& 92.80 (55.80) \\

\alg\ ($K=4$)
& $0.27 \pm 0.02$ & 0.28 & 0.24
& $+3.59$ & $-3.65$
& 93.42
& 90.62
& 92.35 (54.95) \\
\bottomrule
\end{tabular}
}
\end{table*}

Longer federated training substantially reduces the effect of group imbalance. Under the 200-round budget of Section~\ref{sec:motivation}, FL reaches only $\mathrm{acc}^{10}=61.75$ under the imbalanced setting (Table~\ref{tab:imbalanced}), whereas extending training to 400 rounds improves it to $90.55$, indicating that group imbalance slows rather than prevents federated optimization. Nevertheless, \alg still achieves the highest accuracy after only $R/2=200$ rounds of federated training. On the synthetic datasets, FL eventually catches up after sufficient personalization ($93.75$ vs.\ $93.80$ under imbalance and $94.30$ vs.\ $94.70$ under balanced groups), indicating that the benefit of the debiased initialization lies in faster adaptation rather than a better final optimum.

\alg outperforms the baselines across larger models (Table~\ref{tab:realworld}).
On both Qwen 1.5B~\cite{yang2024qwen2technicalreport} and Gemma-2B~\cite{gemmateam2024gemmaopenmodelsbased}, \alg achieves the best
post-adaptation accuracy, and the gain comes from personalization rather than
from a stronger starting point. On Gemma-2B, \alg improves from $70.29$ to
$72.35$ while starting below FL, which gains only $0.18$ points
($71.29 \rightarrow 71.47$). On Qwen-1.5B, \alg improves by $1.64$ points
against FL's $0.65$.

\subsection{Robustness to the Choice of $K$}

\alg does not require identifying the true preference groups.
Section~\ref{sec:motivation} showed that the degradation under imbalance
originates from biased client sampling rather than an incorrect partition.
Accordingly, \alg only requires a partition that removes the sampling
bias, rather than exactly recovering the preference groups.

\begin{figure}
    \centering
    \includegraphics[width=0.85\linewidth]{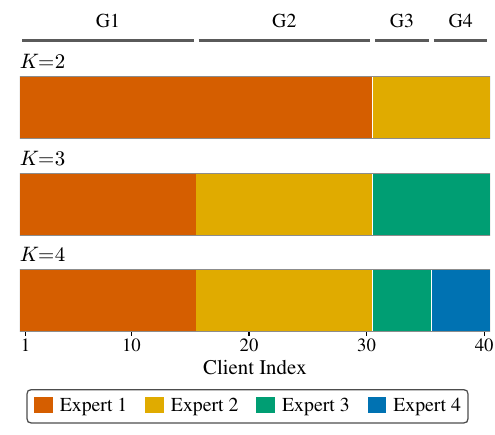}
    \caption{Phase~1 clustering results under the synthetic imbalance setting for different numbers of experts ($K$). Clients are colored by their assigned expert (lowest validation loss).}
    \label{fig:clusters}
\end{figure}

Figure~\ref{fig:clusters} illustrates this mechanism. With ground-truth
groups G1 (Elementary \& Humorous), G2 (Elementary \& Non-Humorous),
G3 (PhD \& Humorous), and G4 (PhD \& Non-Humorous), \alg forms the
partitions $\{G1,G2\}$ and $\{G3,G4\}$ for $K=2$, preserving the shared
education attribute while balancing the humor attribute within each
partition. Increasing $K$ to three further separates G1 and G2, resulting
in the partitions $\{G1\}$, $\{G2\}$, and $\{G3,G4\}$. This preserves both
education and humor information for the Elementary clients, while the
PhD partition continues to average over the humor attribute. Only $K=4$
produces the partitions $\{G1\}$, $\{G2\}$, $\{G3\}$, and $\{G4\}$,
preserving both attributes for every group.

Despite these different partitions, Table~\ref{tab:kdiag} demonstrates that the results retain robust optimization. Across all three values of $K$, the learned models maintain low curvature ($GQ(l)\le0.29$), remain near the decision boundary with narrow initial margins, and achieve high correction rates just 10 local steps. As a result, the accuracy varies marginally, within a range of $91.0\%$ to $ 92.8\%$.

These results indicate that recovering the exact preference groups is
not necessary. Once the discovered partition sufficiently removes the biased client sampling, the resulting initialization retains the flatness and under-confident margins enabling rapid local adaptation, making \alg
largely insensitive to the choice of $K$.

\section{Related Work}
\label{sec:related}

\noindent\textbf{Reward modeling under preference heterogeneity.}\quad
Reward models trained on preference data are central to alignment pipelines \citep{stiennon2020learning, ouyang2022training}, alongside alternatives such as DPO~\citep{rafailov2023direct} and RRHF~\citep{yuan2023rrhf}. To accommodate diverse users, recent methods absorb the variation into the structure of one centralized model, including group-wise robust objectives~\citep{ramesh2024group}, multi-objective reward heads~\citep{wang2024arithmetic}, routing over latent subgroups~\citep{shen2025micro}, latent user variables~\citep{poddar2024personalizing}, and post-hoc merging of objective-specific policies~\citep{jang2023personalized}, so that a single deployed model can be steered per user. All of these approaches assume centralized access to preference data, which privacy regulation often precludes~\citep{regulation2016regulation, illman2019california, kopf2023openassistant}. Federated preference alignment removes this assumption by aggregating locally computed alignment updates~\citep{ye2024openfedllm, wu2024towards, fan2024fedrlhf}, but returning a single consensus policy limits its ability to serve heterogeneous clients, underscoring the need to leverage federated learning for local personalization.

\noindent\textbf{Multiple global models under heterogeneity.}\quad
When clients disagree in preference signals, a common method is to train multiple global models, one per group of similar clients. FedBiscuit~\citep{wu2024towards} clusters clients by preference similarity and trains one reward model per cluster, the design our experiments compare against. FedLEASE~\citep{wang2025adaptive}, IFCA~\citep{ghosh2020efficient}, CFL~\citep{sattler2020clustered}, and FeSEM~\citep{long2023multi} likewise partition clients by preference or domain similarity, while soft membership~\citep{ruan2022fedsoft} and mixture-of-experts approaches~\citep{yi2024pfedmoe} offer more flexible combinations. These designs rest on the premise that a client is better served by a model that never averaged over conflicting objectives. A complementary line of personalized federated learning instead keeps one global model. It defers personalization to local adaptation, either by training the global model as an initialization for a few local gradient steps~\citep{fallah2020personalized, dinh2020personalized, oh2021fedbabu} or by sharing a common representation while keeping heads or lightweight adapters client-specific~\citep{collins2021exploiting, li2021ditto, yi2023pfedlora, scott2024pefll}, and biased client participation in this setting can be corrected through clustered or importance-based sampling~\citep{fraboni2021clustered, chen2022optimal}. However, prior work evaluates models primarily at the global level, so a deeper analysis is needed to examine performance trends after local fine-tuning, thereby clarifying the true role of client grouping.

\vspace{-7pt}
\section{Conclusion}
\label{sec:conclusion}
We examined what constitutes a good initialization for personalized reward modeling under conflicting preferences. The prevailing response to such conflict is to partition clients and train one model per group, on the premise that a client is better served by a model that never averaged over opposing labels. Our results do not support this premise. A single federated model performs at chance before adaptation, yet a few local steps carry it past a centralized model trained on the client's own ground-truth group. Averaging cancels the opposing label choices while retaining the distinction clients agree on, leaving the initialization in a flat region from which successive local steps remain aligned. When the preference groups differ in size the cancellation becomes asymmetric: the shared model commits to the majority side and leaves minority clients too far from the decision boundary for a few local steps to recover. \alg restores the symmetry by discovering the groups during federated training and debiasing the client sampling over them.

\bibliographystyle{Styles/ACM-Reference-Format}
\balance
\bibliography{Styles/kdd}

\appendix

\newpage
\appendix
\supptitle

\appendix

This appendix provides additional details omitted from the main paper.
Appendix~\ref{appsec:exp_setup} describes the experimental setup,
Appendix~\ref{appsec:alg_details} presents the complete training procedures
of \alg\ and the compared methods, and
Appendix~\ref{appsec:additional_results} provides additional experimental
analyses.

\section{Experimental Setup}
\label{appsec:exp_setup}

\subsection{Computing Environment}

Synthetic experiments are conducted on NVIDIA RTX~3090 and RTX~A5000 GPUs
using Python~3.9, PyTorch~2.1.0, and CUDA~12.1. Real-world experiments are
performed on NVIDIA RTX~5090 GPUs using Python~3.12, PyTorch~2.12.0, and
CUDA~13.0. Both environments use the same implementation with
Transformers~4.49.0, PEFT~0.17.1, and Accelerate~1.10.1 under bfloat16 mixed
precision.

\subsection{Implementation Details}

Unless explicitly mentioned otherwise, all experiments share the following
implementation.

\begin{itemize}
\setlength{\itemsep}{3pt}

\item \textbf{Training configuration.}
We run 400 communication rounds of federated learning, sampling five clients
per round ($|S_r|=5$; $|G_r|=5$ for group-debiased methods). Each selected
client performs 30 local optimization steps with a batch size of 16 using
AdamW~\cite{loshchilov2018decoupled} with
$(\beta_1,\beta_2)=(0.9,0.95)$ and a constant learning rate of
$1\times10^{-5}$. Client assignments are refreshed every $T=20$
communication rounds for methods with dynamic grouping. Unless otherwise
specified, multi-global methods use $K=4$, and \alg\ uses a mixing
coefficient of $w=0.6$.

\item \textbf{Backbone models and LoRA.}
We use Qwen2-0.5B~\cite{yang2024qwen2technicalreport} for the synthetic experiments. For the
real-world evaluation, we use the Hugging Face checkpoints
\texttt{Qwen/Qwen2.5-1.5B} and
\texttt{google/gemma-2-2b-it}. Across all backbones, LoRA
fine-tuning~\cite{hu2022lora,houlsby2019parameter} is performed with rank
$r=8$, scaling factor $\alpha=16$, and dropout rate 0.05. LoRA adapters are
inserted into \texttt{q\_proj}, \texttt{k\_proj}, \texttt{v\_proj},
\texttt{o\_proj}, \texttt{gate\_proj}, and \texttt{up\_proj}, while all
backbone parameters remain frozen.

\item \textbf{Reward model implementation.}
Our implementation is built upon the official FedBiscuit repository
(\url{https://github.com/HarliWu/FedBiscuit}). We directly reuse its
binary-selector reward model and training pipeline without architectural
modification, interpreting the resulting two logits as the reward scores
$r_\theta(x,y^{+})$ and $r_\theta(x,y^{-})$ throughout the paper.

\item \textbf{Baseline implementations.}

\begin{itemize}
\setlength{\itemsep}{2pt}

\item \textbf{CL.}
A single reward model is trained on the centralized dataset.

\item \textbf{FL.}
Standard FedAvg~\cite{mcmahan2017communication} is used to optimize a single global reward model.

\item \textbf{Multi-CL.}
Oracle multi-global training is performed using the ground-truth preference
groups, where one centralized reward model is trained for each group.

\item \textbf{Local.}
Each client independently fine-tunes a LoRA adapter without any global
training or communication.

\item \textbf{FedBiscuit.}
The server maintains multiple cluster-specific global models and periodically
reassigns each client to the model with the lowest validation loss.

\item \textbf{Soft-FL.}
Each client receives a validation-weighted fusion of all global models and
contributes to every global model according to the same validation-based
weights.

\end{itemize}

The complete training procedures of \alg, FedBiscuit, and Soft-FL are
provided in Appendix~\ref{appsec:alg_details}.

\end{itemize}

\newpage

\section{Algorithmic Details}
\label{appsec:alg_details}
This appendix presents the complete pseudocode for \alg\ together with the
FedBiscuit~\cite{wu2024towards} and Soft-FL baselines used in our
experiments. Section~\ref{app:fedgd} details the proposed method, while
Sections~\ref{app:alg_fedbiscuit} and~\ref{app:alg_softfusion} describe the
FedBiscuit and Soft-FL implementations used for comparison.

\subsection{FedGD}
\label{app:fedgd}

\begin{algorithm}[H]
\SetAlgoLined
\small
\setstretch{0.9}
\caption{FedGD: Federated Learning with Group Debiasing and Final Personalization}
\label{alg:fedgd}
\KwIn{experts $K$; total rounds $R$; reassignment period $T$; local steps $\tau$; sampling size $|G|$; mixing coefficient $w$; client set $[C]$; dataset sizes $\{N_c\}_{c=1}^{C}$}
\KwResult{single reward model $\phi^{(R)}$ and personalized models $\{\theta^{\text{PFL}}_c\}_{c=1}^C$}
\textbf{Init:} $\theta_g^{(0)}, \{\theta_k^{(0)}\}_{k=1}^K \leftarrow$ LoRA default init;\ 
$\pi(c) \leftarrow \operatorname{Uniform}\{1,\ldots,K\}\ \forall c$;\ 
$\mathcal{A}_k \leftarrow \{c : \pi(c) = k\}\ \forall k$\;
\textbf{Phase 1: Discovering the Groups ($r = 0, \ldots, R/2 - 1$)}\;
\For{$r = 0, 1, \ldots, R/2 - 1$}{
    \If(\tcp*[f]{reassign clients and reset experts}){$r > 0$ and $r \bmod T \equiv 0$}{
        $\pi(c) \leftarrow \arg\min_k \operatorname{ValLoss}(\theta_k^{(r-1)}, D^c_\text{val})\ \forall c$\;
        $\mathcal{A}_k \leftarrow \{c : \pi(c) = k\}\ \forall k$;\ 
        $W_k \leftarrow 0,\ n_k \leftarrow 0\ \forall k$\;
        $\theta_k^{(r-1)} \leftarrow \theta_g^{(r-1)}\ \forall k$
            \tcp*[r]{reset experts to reference}
    }
    $G_r \leftarrow \operatorname{GroupDebiasedSample}(|G|, \{\mathcal{A}_k\})$\;
    \ForEach{$c \in G_r$ in parallel}{
        $\theta^{(r)}_{\pi(c), c} \leftarrow \operatorname{LocalTrain}(\theta_{\pi(c)}^{(r-1)}; D^c_\text{train})$\;
    }
    \tcp{expert update: cumulative average + moving average}
    \For{$k = 1, \ldots, K$}{
        $G_{r,k} \leftarrow G_r \cap \mathcal{A}_k$; \lIf{$G_{r,k} = \varnothing$}{\textbf{continue}}
        $\bar{\theta}_k^{(r)} \leftarrow \tfrac{1}{|G_{r,k}|} \sum_{c \in G_{r,k}} \theta^{(r)}_{k,c}$\;
        $W_k \leftarrow W_k + |G_{r,k}| \cdot \bar{\theta}_k^{(r)}$;\ 
        $n_k \leftarrow n_k + |G_{r,k}|$;\ 
        $s_k \leftarrow W_k / n_k$\;
        $\theta_k^{(r)} \leftarrow (1-w)\, \theta_k^{(r-1)} + w\, s_k$\;
    }
    $\theta_g^{(r)} \leftarrow \tfrac{1}{|G_r|} \sum_{c \in G_r} \theta^{(r)}_{\pi(c), c}$
        \tcp*[r]{reference: uniform average}
}
\tcp{freeze the discovered groups at the end of Phase 1}
$\pi(c) \leftarrow \arg\min_k \operatorname{ValLoss}(\theta_k^{(R/2 - 1)}, D^c_\text{val})\ \forall c$\;
$\mathcal{A}^{\star}_k \leftarrow \{c : \pi(c) = k\}\ \forall k$;\ 
discard empty clusters and reindex $k = 1, \ldots, K$\;
\textbf{Phase 2: Training the Reward Model ($r = R/2, \ldots, R - 1$)}\;
Initialize $\phi^{(R/2)} \leftarrow$ LoRA default init\;
\For{$r = R/2, R/2 + 1, \ldots, R - 1$}{
    $G_r \leftarrow \operatorname{GroupDebiasedSample}(|G|, \{\mathcal{A}^{\star}_k\})$;\ 
    broadcast $\phi^{(r)}$ to $G_r$\;
    \ForEach{$c \in G_r$ in parallel}{
        $\phi^{(r+1)}_c \leftarrow \operatorname{LocalTrain}(\phi^{(r)}; D^c_\text{train})$\;
    }
    \tcp{hierarchical: size-weighted within a group, uniform across groups}
    $G^{\star}_{r,k} \leftarrow G_r \cap \mathcal{A}^{\star}_k\ \forall k$;\ 
    $\mathcal{K}_r \leftarrow \{k : G^{\star}_{r,k} \neq \varnothing\}$\;
    \For{$k \in \mathcal{K}_r$}{
        $\bar{\phi}_k^{(r+1)} \leftarrow
          \sum_{c \in G^{\star}_{r,k}}
          \frac{N_c}{\sum_{c' \in G^{\star}_{r,k}} N_{c'}}\, \phi^{(r+1)}_c$\;
    }
    $\phi^{(r+1)} \leftarrow
      \frac{1}{|\mathcal{K}_r|} \sum_{k \in \mathcal{K}_r} \bar{\phi}_k^{(r+1)}$\;
}
\textbf{Final Personalization}\;
Server broadcasts $\phi^{(R)}$ to all clients\;
\ForEach{$c \in [C]$ in parallel}{
    $\theta^{(\text{PFL}, 0)}_c \leftarrow \phi^{(R)}$; fine-tune on $D^c_\text{train}$ to obtain $\theta^{\text{PFL}}_c$\;
}
\Return $\phi^{(R)},\ \{\theta^{\text{PFL}}_c\}_{c=1}^C$\;
\end{algorithm}

\newpage
\paragraph{Description.}
Algorithm~\ref{alg:fedgd} implements \alg, which applies group debiasing when the
true preference groups are unknown. Unlike the multi-global designs of
FedBiscuit~\cite{wu2024towards}, the $K$ expert models are not the object of training:
they exist only to \emph{discover} a partition of the clients, and the model that
clients ultimately personalize is a single reward model $\phi$ trained afterwards.
Like the single-global design of Section~\ref{subsec:fl-procedure}, each client
receives exactly one model per round, so the per-round communication cost is
matched.

\textbf{Phase 1 (rounds $0$ to $R/2-1$).} The server maintains $K$ experts
$\{\theta_k\}_{k=1}^K$ together with a reference model $\theta_g$. Clients start
from a uniform random assignment, and every $T$ rounds each client evaluates all
$K$ experts on its own validation split and joins the one with the lowest
validation loss~\citep{wu2024towards}, giving the groups
$\{\mathcal{A}_k\}_{k=1}^K$. Each reassignment also resets every expert to the
current reference model, $\theta_k \leftarrow \theta_g$, so that an expert does not
carry over what it learned from the clients it held before the reassignment. In the
rounds between two reassignments, the server draws the participating set $G_r$ with
group-debiased sampling over the current groups, sends each selected client the
expert of its own group, and receives the locally trained model. Because a single
round contributes only $|G_r|/K$ clients to a given expert on average, the server
accumulates the returned models since the last reassignment and takes their
cumulative average $s_k$, then updates the expert as a moving average
$\theta_k^{(r)} = (1-w)\theta_k^{(r-1)} + w\,s_k$; this estimates each expert from
more clients than a single round provides and damps the round-to-round variance of
the estimate. The reference model is updated by a uniform average over $G_r$: it
serves as a neutral point to which the experts are reset rather than as a loss
minimizer, and since every client takes the same $\tau$ local steps, weighting by
dataset size here would let the discovered partition follow data volume rather than
preference. At the end of Phase~1 the clients are reassigned once more, empty
clusters are discarded, and the resulting groups $\{\mathcal{A}_k^{\star}\}$ are
frozen.

\textbf{Phase 2 (rounds $R/2$ to $R-1$).} The remaining half of the budget trains a
single reward model $\phi$ from a fresh LoRA initialization, with group-debiased
sampling over the frozen groups and hierarchical aggregation: the returned models
are averaged in proportion to $N_c$ within each group, and the resulting group
models are then averaged uniformly. Debiasing the sampling equalizes how often a
group is heard, and aggregating hierarchically keeps that equality at the server,
where weighting every client by $N_c$ would otherwise let a group with larger
clients dominate a round it shares with a smaller one. Starting from a fresh
initialization rather than from $\theta_g$ isolates $\phi$ from the intermediate,
unstable partitions that Phase~1 passes through; we compare the two choices
empirically in Appendix~\ref{appsec:phase2}. Note that $\phi$ therefore receives only
$R/2$ rounds of training, half the budget given to the single-global and
multi-global baselines.

\textbf{Final personalization.} The server broadcasts $\phi^{(R)}$ to all clients,
and each client fine-tunes it on $D^c_{\text{train}}$ without further communication
to obtain $\theta_c^{\text{PFL}}$. In the notation of
Section~\ref{subsec:fl-procedure}, \alg\ is a single-global design with
$\theta_c^{\text{init}} = \phi^{(R)}$ for every client; the $K$ experts and the
reference model are discarded after Phase~1. When the discovered partition
coincides with the true preference groups, this reduces exactly to the FL\_target
oracle of Section~\ref{sec:motivation-imbalanced}.

\subsection{FedBiscuit~\cite{wu2024towards}}
\label{app:alg_fedbiscuit}
\paragraph{Description.}
Algorithm~\ref{alg:fedbiscuit} is the federated counterpart of Multi-CL: it
trains one global model per preference group while discovering the groups from
decentralized data instead of assuming them. We follow
FedBiscuit~\citep{wu2024towards} and reinterpret it in an EM-style
manner~\citep{dempster1977maximum} for our personalized reward modeling
setting. The server maintains $K$ cluster-specific global models but sends only
one of them to each client per round, so the per-round communication cost
matches the single-global design of
Section~\ref{subsec:fl-procedure}. Cluster assignments are refreshed every
$T$ rounds: the \emph{E-step} assigns each client to the model with the lowest
validation loss on $D^c_{\mathrm{val}}$, and the \emph{M-step} performs $T$
rounds of clustered federated learning, where each cluster updates only its
assigned model.

Phase~1 warms up each model independently with FedAvg for $T$ rounds so that
the first E-step does not operate on $K$ identical models. Without this
warm-up, every client would observe identical validation losses for all
cluster-specific models, making the initial assignment arbitrary. Phase~2 then
alternates between the E-step and M-step. Since assignments based solely on
validation loss may produce highly imbalanced clusters, each reassignment is
followed by the original size-balancing heuristic, which iteratively moves
clients from the largest cluster to the smallest until the cluster sizes differ
by at most one while minimizing the increase in validation loss.

Aggregation follows the original FedBiscuit
rule~\citep{wu2024towards},
\begin{equation}
    \theta_k^{(r)}
    =
    \Bigl(1 - \sum_{c \in S_{r,k}} p_c \Bigr)\,\theta_k^{(r-1)}
    + \sum_{c \in S_{r,k}} p_c \,\theta^{(r)}_{k,c},
    \qquad
    p_c = \frac{N_c}{\sum_{j \in [C]} N_j},
    \label{eq:fedbiscuit_agg}
\end{equation}
where $S_{r,k}$ denotes the sampled clients assigned to model $k$ at round
$r$. Since clients are sampled uniformly from $[C]$ rather than from each
cluster, the total weight $\sum_{c \in S_{r,k}} p_c$ varies substantially
across rounds. Directly averaging only the participating clients would
therefore produce unstable updates when few clients from a cluster are sampled.
The residual term preserves a
$(1-\sum_{c\in S_{r,k}}p_c)$ fraction of the previous model, stabilizing the
optimization under sparse cluster participation.

In Phase~3, each client selects the model among
$\{\theta_k^{(R)}\}_{k=1}^K$ with the lowest validation loss on its validation
split and fine-tunes it locally, yielding
$\theta_c^{\text{init}}=\theta_{\pi_c^{\star}}^{(R)}$ in the notation of
Section~\ref{subsec:fl-procedure}. Unlike \alg, which discards the auxiliary
experts after Phase~1 and deploys a single reward model, FedBiscuit retains all
$K$ cluster-specific models at deployment, requiring each client to evaluate
all $K$ models before personalization.

\begin{algorithm}[H]
\small
\caption{FedBiscuit: Federated Training with Size-Balanced Hard-Clustered Global Models and Final Personalization}
\label{alg:fedbiscuit}
\KwInput{%
  number of global models $K$; total rounds $R$; reassignment period $T$;
  local steps $\tau$; client set $[C]$; dataset sizes $\{N_c\}_{c=1}^{C}$;
  initial parameters $\{\theta_k^{(0)}\}_{k=1}^K$}
\KwResult{%
  clustered global models $\{\theta_k^{(R)}\}_{k=1}^K$ and
  personalized models $\{\theta_c^{\text{PFL}}\}_{c=1}^C$}

\BlankLine
$p_c \leftarrow N_c / \sum_{j \in [C]} N_j \quad \forall c$\;

\BlankLine
\textbf{Phase 1: Warm-up (independent FedAvg for each model)}\;
\For{$k = 1,2,\dots,K$}{
  \For{$t = 1,2,\dots,T$}{
    Sample a client subset $S_{t,k} \subseteq [C]$\;
    \ForEach{$c \in S_{t,k}$ \textbf{in parallel}}{
      Client $c$ receives $\theta_k^{(t-1)}$\;
      Run $\tau$ local update steps on $D_{\text{train}}^c$ and send
      $\theta^{(t)}_{k,c}$ to the server\;
    }
    $\theta_k^{(t)} \leftarrow \frac{1}{|S_{t,k}|}
      \sum_{c \in S_{t,k}} \theta^{(t)}_{k,c}$
      \tcp*[r]{FedAvg update for model $k$}
  }
}

\BlankLine
\textbf{Phase 2: Size-balanced hard assignment}\;
\For{$r = T+1,T+2,\dots,R$}{
  \If{$r-1 \equiv 0\;(\bmod T)$}{
    \tcp{(1) Client-driven reassignment by validation loss}
    Server broadcasts $\{\theta_k^{(r-1)}\}_{k=1}^K$ to all clients\;
    \ForEach{$c \in [C]$}{
      $\ell_{c,k} \gets
        \operatorname{ValLoss}(\theta_k^{(r-1)}, D^c_{\mathrm{val}})
        \ \ \forall k$;\quad
      $\pi(c) \gets \arg\min_{k} \ell_{c,k}$\;
    }
    $\mathcal{A}_k \leftarrow \{\, c : \pi(c) = k \,\}\ \forall k$\;
    \tcp{(2) Size balancing}
    \While{$\max_k |\mathcal{A}_k| - \min_k |\mathcal{A}_k| > 1$}{
      $k^{+} \gets \arg\max_k |\mathcal{A}_k|$;\quad
      $k^{-} \gets \arg\min_k |\mathcal{A}_k|$\;
      $c^{\star} \gets \arg\min_{c \in \mathcal{A}_{k^{+}}}
        \bigl( \ell_{c,k^{-}} - \ell_{c,k^{+}} \bigr)$
        \tcp*[r]{least-cost move}
      $\mathcal{A}_{k^{+}} \leftarrow \mathcal{A}_{k^{+}} \setminus \{c^{\star}\}$;\quad
      $\mathcal{A}_{k^{-}} \leftarrow \mathcal{A}_{k^{-}} \cup \{c^{\star}\}$;\quad
      $\pi(c^{\star}) \leftarrow k^{-}$\;
    }
  }
  \tcp{(3) Uniform sampling and cluster-wise training}
  Sample a subset of participating clients $S_r \subseteq [C]$\;
  \For{$k = 1,2,\dots,K$}{
    $S_{r,k} \leftarrow S_r \cap \mathcal{A}_k$
      \tcp*[r]{$\{S_{r,k}\}_{k=1}^K$ partitions $S_r$}
    \If{$S_{r,k} \neq \emptyset$}{
      Server sends $\theta_k^{(r-1)}$ to all $c \in S_{r,k}$\;
      \ForEach{$c \in S_{r,k}$ \textbf{in parallel}}{
        Client $c$ runs $\tau$ local update steps on $D_{\text{train}}^c$
        and sends $\theta^{(r)}_{k,c}$ to the server\;
      }
      $\theta_k^{(r)} \leftarrow
        \bigl(1 - \textstyle\sum_{c \in S_{r,k}} p_c \bigr)\theta_k^{(r-1)}
        + \sum_{c \in S_{r,k}} p_c \,\theta^{(r)}_{k,c}$
        \tcp*[r]{weighted aggregation with residual}
    }
  }
}

\BlankLine
\textbf{Phase 3: Final personalization}\;
Server broadcasts $\{\theta_k^{(R)}\}_{k=1}^K$ to all clients\;
\ForEach{$c \in [C]$}{
  $\pi_c^{\star} \gets \arg\min_{k}
    \operatorname{ValLoss}(\theta_k^{(R)}, D^c_{\mathrm{val}})$\;
}
\ForEach{$c \in [C]$ \textbf{in parallel}}{
  $\theta_c^{(\text{PFL},0)} \leftarrow \theta_{\pi_c^{\star}}^{(R)}$;
  fine-tune on $D^c_{\text{train}}$ to obtain $\theta_c^{\mathrm{PFL}}$\;
}
\Return{$\{\theta_k^{(R)}\}_{k=1}^K,\ \{\theta_c^{\text{PFL}}\}_{c=1}^C$}\;
\end{algorithm}

\newpage
\subsection{Soft-Clustered Multi-Global Models (Soft-FL)}
\label{app:alg_softfusion}

\begin{algorithm}[ht!]
\small
\caption{Soft-FL: Federated Training with Soft-Clustered Global Models and Final Personalization}
\label{alg:softfusion}
\SetKwFunction{UpdateW}{UpdateSoftWeights}
\KwInput{%
  number of global models $K$; total rounds $R$; refresh period $T$;
  local steps $\tau$; client set $[C]$;
  initial global parameters $\{\theta_k^{(0)}\}_{k=1}^K$}
\KwResult{%
  global models $\{\theta_k^{(R)}\}_{k=1}^K$ and
  personalized models $\{\theta_c^{\text{PFL}}\}_{c=1}^C$}

\BlankLine
$\{w_{k,c}^{(0)}\}_{k,c} \leftarrow$ \UpdateW{$\{\theta_k^{(0)}\}_{k=1}^K$}
  \tcp*[r]{initial soft assignments}

\BlankLine
\textbf{Phase 1: Federated multi-global training}\;
\For{$r = 1,2,\dots,R$}{
  Sample participating clients $S_r \subseteq [C]$\;
  \tcp{(1) Server $\rightarrow$ clients: fused initialization}
  \ForEach{$c \in S_r$}{
    $\theta^{\mathrm{fuse},(r-1)}_c \leftarrow
      \sum_{k=1}^K w_{k,c}^{(r-1)} \theta_k^{(r-1)}$;\quad
    send $\theta^{\mathrm{fuse},(r-1)}_c$ to client $c$\;
  }
  \tcp{(2) Local training at clients}
  \ForEach{$c \in S_r$ \textbf{in parallel}}{
    Run $\tau$ local update steps on $D^c_{\text{train}}$ from
    $\theta^{\mathrm{fuse},(r-1)}_c$ and send $\theta_c^{+,(r)}$ to the server\;
  }
  \tcp{(3) Expert-wise aggregation at the server}
  \For{$k = 1,2,\dots,K$}{
    $\theta_k^{(r)} \leftarrow
      \frac{1}{Z_k^{(r)}} \sum_{c \in S_r} w_{k,c}^{(r-1)}\, \theta_c^{+,(r)}$,
    \quad $Z_k^{(r)} = \sum_{c \in S_r} w_{k,c}^{(r-1)}$\;
  }
  \tcp{(4) Refresh soft weights every $T$ rounds and at $r{=}R$}
  \If{$(r \bmod T = 0)\ \mathrm{or}\ r = R$}{
    $\{w_{k,c}^{(r)}\}_{k,c} \leftarrow$ \UpdateW{$\{\theta_k^{(r)}\}_{k=1}^K$}\;
  }
}

\BlankLine
\textbf{Phase 2: Final personalization}\;
\ForEach{$c \in [C]$}{
  $\theta_c^{\text{init}} \leftarrow \sum_{k=1}^K w_{k,c}^{(R)} \theta_k^{(R)}$;\quad
  send $\theta_c^{\text{init}}$ to client $c$\;
}
\ForEach{$c \in [C]$ \textbf{in parallel}}{
  $\theta_c^{(\text{PFL},0)} \leftarrow \theta_c^{\text{init}}$;
  fine-tune on $D^c_{\text{train}}$ to obtain $\theta_c^{\mathrm{PFL}}$\;
}

\BlankLine
\SetKwProg{Fn}{Function}{:}{}
\Fn{\UpdateW{$\{\theta_k\}_{k=1}^K$}}{
  Server broadcasts $\{\theta_k\}_{k=1}^K$ to all clients\;
  \ForEach{$c \in [C]$ \textbf{in parallel}}{
    $a_{k,c} \gets \operatorname{ValAcc}(\theta_k, D^c_{\mathrm{val}})
      \ \ \forall k$\;
    $w_{k,c} \leftarrow a_{k,c} / \sum_{j=1}^{K} a_{j,c}
      \ \ \forall k$;\quad
    send $\{w_{k,c}\}_{k=1}^K$ to the server\;
  }
  \Return{$\{w_{k,c}\}_{k \in [K],\, c \in [C]}$}\;
}
\Return{$\{\theta_k^{(R)}\}_{k=1}^K,\ \{\theta_c^{\text{PFL}}\}_{c=1}^C$}\;
\end{algorithm}

\paragraph{Description.}
Algorithm~\ref{alg:softfusion} replaces the hard cluster assignment of
FedBiscuit~(Algorithm~\ref{alg:fedbiscuit}) with a soft one. Each client $c$ holds
a weight vector $\{w_{k,c}\}_{k=1}^K$ over the $K$ global models, obtained by
normalizing its validation accuracies, so every model is updated by every client
in proportion to how well it fits that client. This is a mixture-of-experts style
soft assignment~\citep{jacobs1991adaptive, shazeer2017outrageously}, but the
communication pattern is unchanged: a client still receives exactly one model per
round, namely the \emph{fused} model
$\theta^{\mathrm{fuse},(r-1)}_c = \sum_{k} w_{k,c}^{(r-1)} \theta_k^{(r-1)}$
formed at the server.

After local training the client returns a single model $\theta_c^{+,(r)}$, from
which all $K$ experts must be updated. We assign it to each expert with that
client's weight, which amounts to solving
\[
    \theta_k^{(r)}
    = \arg\min_{\theta} \sum_{c \in S_r} w_{k,c}^{(r-1)}
      \bigl\lVert \theta - \theta_c^{+,(r)} \bigr\rVert_2^2 ,
\]
whose closed form is the weighted average in Algorithm~\ref{alg:softfusion}. The
weights are refreshed every $T$ rounds by broadcasting all $K$ models for
validation, the soft counterpart of the E-step in FedBiscuit. In Phase~2, client
$c$ is initialized from its own fused model
$\theta_c^{\text{init}} = \sum_{k} w_{k,c}^{(R)} \theta_k^{(R)}$ and fine-tunes it
locally without further communication.

\smallskip
\noindent\textbf{Adapter initialization.}\quad
Soft-FL requires the $K$ models to differ at $r = 0$, which the standard LoRA
initialization does not provide. There, the down-projection $A$ is random and the
up-projection $B$ is zero, so $BA = 0$ and every model is functionally identical to
the frozen base model; all validation accuracies coincide, the weights $w_{k,c}$
are uniform in $k$, and both the fusion and the expert-wise aggregation collapse to
the single-global design of Section~\ref{subsec:fl-procedure}. We cannot break this
tie with a warm-up phase as FedBiscuit does, since warming up $K$ models for $T$
rounds each would cost $TK$ additional rounds and break the matched communication
budget.

We instead initialize the $K$ adapters as small, mutually orthogonal perturbations
of the base weights. For a LoRA module with output dimension $d_{\mathrm{out}}$,
input dimension $d_{\mathrm{in}}$, and rank $r$, we draw Gaussian random matrices
and take their QR decompositions to obtain orthonormal
$U_{\mathrm{big}} \in \mathbb{R}^{d_{\mathrm{out}} \times Kr}$ and
$V_{\mathrm{big}} \in \mathbb{R}^{d_{\mathrm{in}} \times Kr}$. Adapter $k$ takes
the $k$-th contiguous column block $U_k, V_k$ and is set to
\[
  A_k \gets V_k^\top,
  \qquad
  B_k \gets \varepsilon_{\text{layer}}\,(1 + \text{jitter} \cdot k)\, U_k ,
\]
so that the $K$ updates $B_k A_k$ span mutually orthogonal subspaces. Here
$\varepsilon_{\text{layer}}$ scales with the Frobenius norm of the corresponding
base weight (a fan-in heuristic when unavailable) and $\text{jitter}$ adds a small
per-adapter offset; we use $\varepsilon_{\text{layer}} = 0.005$ and
$\text{jitter} = 0.01$, and apply the same procedure to LoRA embedding modules when
present. Each model is thus a slightly different perturbation of the base model,
which makes the initial validation accuracies and hence the soft weights
non-uniform while keeping all updates in a small neighborhood of the base weights.

\newpage
\section{Additional Experimental Results}
\label{appsec:additional_results}

This section provides additional experimental analyses that complement the
main results. We first evaluate the sensitivity of \alg\ to the number of
experts. We then analyze the optimization behavior of FL under different
local optimization budgets and learning rates. Next, we evaluate the effect
of the Phase~2 initialization strategy.

\subsection{Sensitivity to the Number of Experts}
\label{appsec:k_sensitivity}

Table~\ref{tab:cluster_full} reports the personalized performance under
different numbers of experts ($K\in\{2,3,4\}$). Across both the imbalanced
synthetic and real-world datasets, \alg\ consistently maintains stable
performance across different choices of $K$, indicating that its performance
is not sensitive to the exact number of experts.

\begin{table}[t]
\centering
\small
\renewcommand{\arraystretch}{1.15}
\caption{Effect of the number of clusters $K$ on personalized accuracy,
averaged over clients.}
\label{tab:cluster_full}
\vspace{-8pt}
\begin{tabular*}{0.7\columnwidth}{@{} l @{\extracolsep{\fill}} ccc @{\hspace{8pt}} ccc @{}}
\toprule
& \multicolumn{3}{c}{Imbalanced ($\mathrm{acc}^{80}$)}
& \multicolumn{3}{c}{Real-world ($\mathrm{acc}^{240}$)} \\
\cmidrule(lr){2-4} \cmidrule(lr){5-7}
Method & $2$ & $3$ & $4$ & $2$ & $3$ & $4$ \\
\midrule
FedBiscuit   & 73.50 & 76.95 & 80.80 & 58.94 & 57.52 & 57.06 \\
Soft-FL      & 84.45 & 86.30 & 86.20 & 57.23 & 56.77 & 57.00 \\
\addlinespace[2pt]
\alg (ours)  & \textbf{93.50} & \textbf{93.65} & \textbf{93.75}
             & \textbf{64.47} & \textbf{64.77} & \textbf{64.41} \\
\bottomrule
\end{tabular*}
\end{table}

\subsection{Optimization Sensitivity}
\label{appsec:sensitivity}

\begin{figure}[t]
    \centering
    \includegraphics[width=0.58\columnwidth]{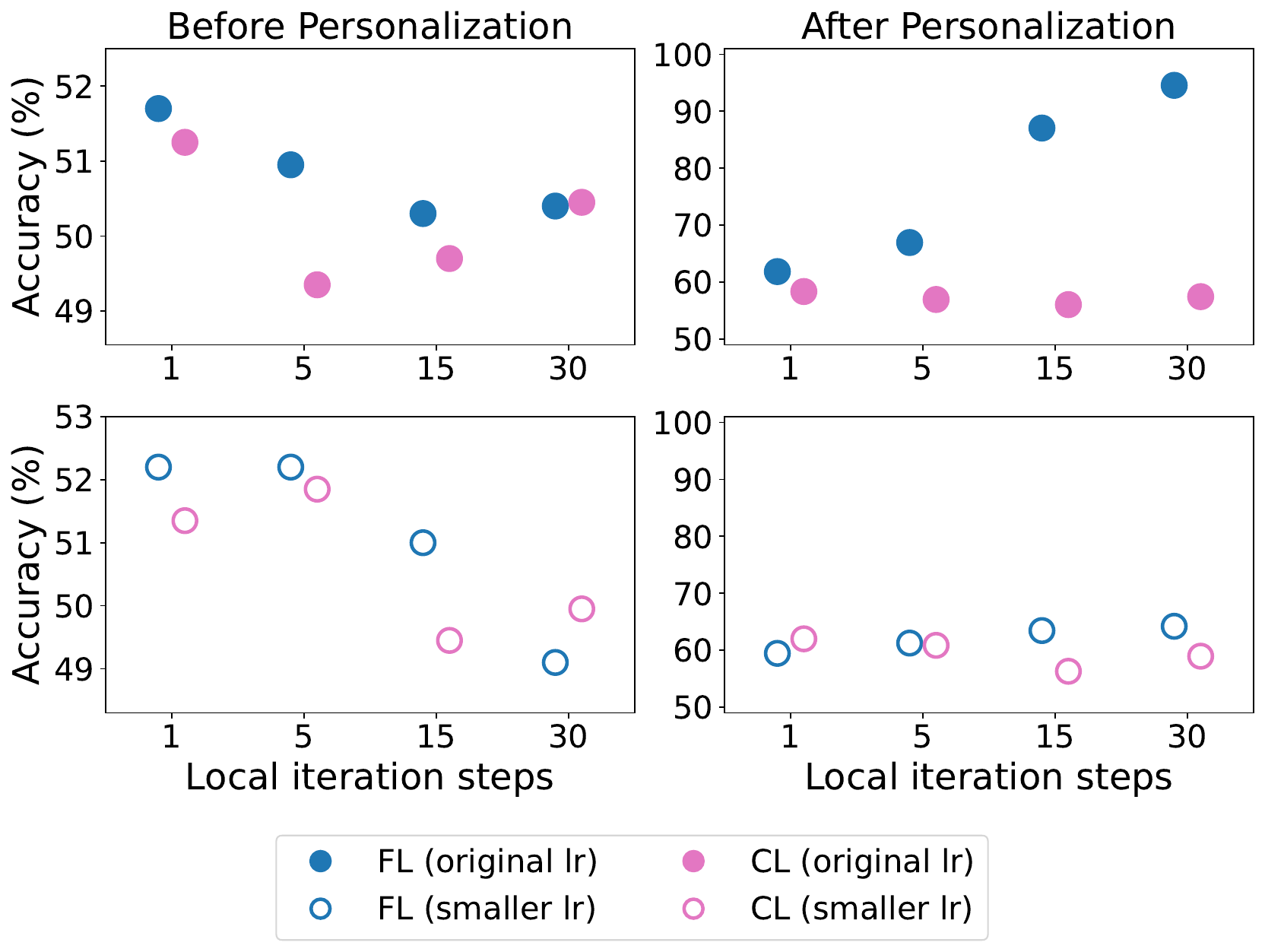}
    \caption{Sensitivity to the local optimization budget and learning rate on
    the balanced synthetic dataset. We compare FL and CL by
    varying the number of local optimization steps before aggregation
    ($\tau\in\{1,5,15,30\}$) and using either the default learning rate or a
    learning rate that is $10\times$ smaller. The left and right columns
    report the global accuracy before and after personalization,
    respectively.}
    \label{fig:single-centralize-sensitivity}
\end{figure}

Figure~\ref{fig:single-centralize-sensitivity} further studies the effect of
the local optimization budget and learning rate on FL and centralized
learning (CL). The top row shows the results with the default learning rate,
whereas the bottom row uses a learning rate that is $10\times$ smaller. The
left and right columns report the global accuracy before personalization
($\mathrm{acc}^{0}$) and the personalized accuracy after 10 local adaptation
steps ($\mathrm{acc}^{10}$), respectively. We vary the number of local
optimization steps before aggregation ($\tau\in\{1,5,15,30\}$).

With the default learning rate, FL and CL achieve similar global accuracies
before personalization, whereas FL yields increasingly higher personalized
accuracies after 10 local adaptation steps as the local optimization budget
grows. In contrast, this trend is largely diminished with the smaller
learning rate, suggesting that sufficient local adaptation before
aggregation, rather than simply increasing the optimization budget, is
essential for obtaining a better initialization for personalization.

\newpage

\subsection{Phase 2 Initialization}
\label{appsec:phase2}

\begin{table}[h]
\centering
\caption{Effect of the Phase~2 initialization strategy for the target reward
model $\phi$. \emph{Continue} warm-starts $\phi$ from the Phase~1 global model
$\theta_g$, whereas \emph{Random} initializes only $\phi$ from scratch while
preserving the expert assignments obtained in Phase~1. Entries report
personalized accuracies (\%) averaged over clients, where
$\mathrm{acc}^{t}$ denotes the accuracy after $t$ local fine-tuning steps.
Bold indicates the better result in each column.}
\label{tab:phase2-init}
\resizebox{0.55\columnwidth}{!}{%
\begin{tabular}{lccc|ccc}
\toprule
& \multicolumn{3}{c|}{Imbalanced} & \multicolumn{3}{c}{Balanced} \\
\cmidrule(lr){2-4}\cmidrule(lr){5-7}
Phase 2 init & $\mathrm{acc}^{0}$ & $\mathrm{acc}^{10}$ & $\mathrm{acc}^{80}$
             & $\mathrm{acc}^{0}$ & $\mathrm{acc}^{10}$ & $\mathrm{acc}^{80}$ \\
\midrule
Continue      & 50.20 & 91.00 & 92.45 & 49.00 & 87.95 & 90.10 \\
Random (ours) & 54.95 & \textbf{92.35} & \textbf{93.75} & 49.15 & \textbf{94.55} & \textbf{94.30} \\
\bottomrule
\end{tabular}%
}
\end{table}

During Phase 1, the client-to-expert assignments gradually converge to the
underlying preference groups, but they remain noisy for a substantial number of
rounds before stabilizing. The reference model $\theta_g$ is updated throughout
this period, so it may already encode optimization bias accumulated under
unreliable partitions. To evaluate this effect, we compare two initialization
strategies for the target reward model $\phi$ in Phase 2: \emph{Continue}, which
warm-starts $\phi$ from $\theta_g$, and \emph{Random}, which initializes only
$\phi$ from scratch while preserving the expert assignments learned in Phase 1.
As shown in Table~\ref{tab:phase2-init}, random initialization consistently
achieves higher personalized accuracy after local adaptation.

\begin{figure}[h]
    \centering
    \includegraphics[width=0.57\linewidth]{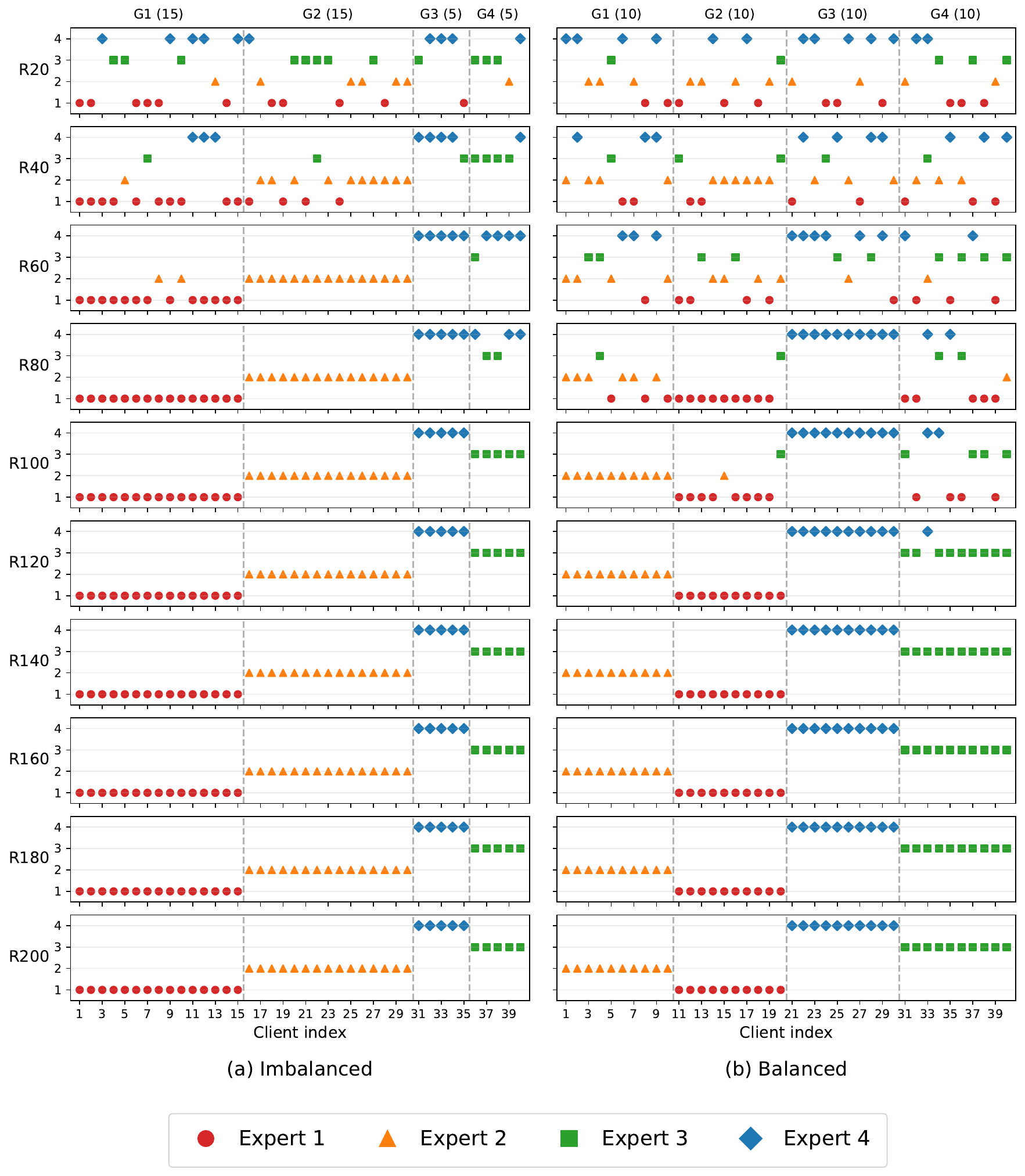}
    \caption{Evolution of client-to-expert assignments during Phase~1 on the
    synthetic dataset. Each marker denotes the expert assigned to a client at
    the corresponding communication round. The left and right columns show
    the imbalanced and balanced settings, respectively.}
    \label{fig:expert_assignment_synthetic}
\end{figure}
\newpage
Figure~\ref{fig:expert_assignment_synthetic} explains why this gap is wider under balanced
groups. The assignments settle by round 100 in the imbalanced setting, whereas
the balanced setting continues to reassign clients until round 140. $\theta_g$ is
therefore updated under unreliable partitions for a longer portion of Phase 1 in
the balanced setting, and warm-starting $\phi$ from it costs 6.60 points of
$\mathrm{acc}^{10}$ (87.95 vs.\ 94.55) against 1.35 points under imbalance.
Random initialization removes this inherited bias while retaining the converged
grouping structure, and we adopt it for Phase 2 in all experiments.

\newpage

\end{document}